\documentclass[useAMS,referee,usenatbib,nomarkers,accents ]{article}

\usepackage{amsmath,amssymb,natbib,hyperref,graphicx,xcolor}
\usepackage{amsthm}
\usepackage[margin=1.5in]{geometry}

\newcommand{\bD}{ {\bf D} }
\newcommand{\bE}{ {\bf E} }
\newcommand{\bI}{ {\bf I} }
\newcommand{\bX}{ {\bf X} }
\newcommand{\bY}{ {\bf Y} }
\newcommand{\bZ}{ {\bf Z} }
\newcommand{\bG}{ {\bf G} }
\newcommand{\bC}{ {\bf C} }
\newcommand{\bU}{ {\bf U} }
\newcommand{\bR}{ {\bf R} }
\newcommand{\bS}{ {\bf S} }

\newcommand{\bW}{ {\bf W} }
\newcommand{\bV}{ {\bf V} }

\newcommand{\bA}{ {\bf A} }

\newcommand{\bzero}{ {\bf 0} }

\newcommand{\bThet}{ \mbox{\boldmath $ \Theta $} }

\newtheorem{prop}{Proposition} 
\newtheorem{theorem}{Theorem}
\newtheorem{definition}{Definition}
\newtheorem{lemma}{Lemma}

\begin{document}

\title{Integrative Factorization of Bidimensionally Linked Matrices}
\author{Jun Young Park and Eric F. Lock \\ \\
Division of Biostatistics \\ University of Minnesota School of Public Health \\ 
Minneapolis, Minnesota, U.S.A. }

\date{}

\maketitle

\begin{abstract}
Advances in molecular ``omics'' technologies have motivated new methodology for the integration of multiple sources of high-content biomedical data.  However, most statistical methods for integrating multiple data matrices only consider data shared {\it vertically} (one cohort on multiple platforms) or {\it horizontally} (different cohorts on a single platform).  This is limiting for data that take the form of {\it bidimensionally linked} matrices (e.g., multiple cohorts measured on multiple platforms), which are increasingly common in large-scale biomedical studies. In this paper, we propose BIDIFAC (Bidimensional Integrative Factorization) for integrative dimension reduction and signal approximation of bidimensionally linked data matrices.  Our method factorizes the data into (i) globally shared, (ii) row-shared, (iii) column-shared, and (iv) single-matrix structural components, facilitating the investigation of shared and unique patterns of variability.   For estimation we use a penalized objective function that extends the nuclear norm penalization for a single matrix.  As an alternative to the complicated rank selection problem, we use results from random matrix theory to choose tuning parameters. We apply our method to integrate two genomics platforms (mRNA and miRNA expression) across two sample cohorts (tumor samples and normal tissue samples) using the breast cancer data from TCGA.   We provide R codes for fitting BIDIFAC, imputing missing values, and generating simulated data. \\

\noindent \emph{Keywords}:
BIDIFAC; Bidimensional data; Data integration; Dimension reduction; Principal component analysis
\end{abstract}

\newpage

\section{Introduction}
\label{sec:intro}

\subsection{Overview}
Several recent methodological developments have been motivated  by the integration of multiple sources of genetic, genomic, epigenomic, transcriptomic, proteomic, and other omics data. Successful integration of these disparate but related sources is essential for a complete understanding of the molecular underpinnings of human diseases, by providing essential tools for novel hypotheses \citep{hawkins10} and improving statistical power. For example, TWAS and MetaXcan/PrediXcan have shown improved power for gene-based association testing, by integrating expression quantitative trait loci (eQTL) and genome-wide association (GWAS) data \citep{gamazon15, gusev16}.  Similarly, the iBAG approach \citep{wang2012ibag} has shown improved power for gene-based association by integrating messenger RNA (mRNA) levels with epigenomic data (e.g., DNA methylation).  Moreover, recent studies showed that protein-level variations explain additional individual phenotypic differences not explained by the mRNA levels \citep{wu13}.

In addition to integrating multiple sources of high-dimensional data, integrating high-dimensional data across multiple patient cohorts can also improve interpretation and statistical power. For example,  the integration of genome-wide data from multiple types of cancers can improve classification of oncogenes or tumor suppressors \citep{kumar2015statistically} and may improve clinical prognoses \citep{liu2018integrated}.

 Most statistical methods for the integration of high-dimensional matrices apply to data that are linked {\it vertically} (e.g., one cohort measured with more than one platforms, such as mRNA and miRNA) or {\it horizontally} (e.g., mRNA expression measured for multiple cohorts) \citep{tseng12}. However, linked structures in molecular biomedical data are often more complex.  In particular, the integration of {\it bidimensionally} linked data (e.g., more than one heterogeneous groups of subjects measured by more than one platform) is largely unaddressed.  In this paper, we propose a new statistical method for the  low-rank structural factorization of large bidimensionally linked datasets.  This can be used to accomplish three important tasks: (i) missing value imputation (ii) dimension reduction, and (iii) the interpretation of lower-dimensional patterns that are shared across matrices or unique to particular matrices.  

\subsection{Motivating Example}
\label{example}
The Cancer Genome Atlas (TCGA) is the most comprehensive and well-curated study of the cancer genome, with data for 6 different {\it omics} data sources from 11,000 patients representing 33 different cancer tumor types as well as rich clinical phenotypes.  We consider integrating a cohort of breast cancer (BRCA) tumor samples, and a cohort of normal adjacent tissue (NAT) samples, from TCGA.  NAT samples are often used for differential analyses, e.g., to identify genes with mean differential expression between cancer and normal tissue.  However, such analyses do not address the molecular heterogeneity or trans-omic interactions that characterize cancer cells.  Noticeably, \citet{aran17} conducted a comprehensive study on NAT across different cancers using TCGA and the Genotype-Tissue Expression (GTEx) program data and showed that the expression levels of NAT from breast, colon, liver, lung, and uterine tumors yield different clustering from their respective tumor tissues. In addition, \citet{huang16}, using TCGA data, suggested that NATs not only serve as controls to tumor tissues but also provide useful information on patients' survival that tumor samples do not. More detailed investigations of the molecular heterogeneity between tumor and NAT tissue are limited by available statistical methods, especially for multi-omics data. Our premise is that comprehensive analysis of multiple {\it omics} data sources across both tumor tissues and NAT would distinguish the joint signals that are shared across different {\it omics} profiles (e.g., mRNA and miRNA) and those that are only attributed to the tumors.

\subsection{Existing Methods on Joint Matrix Factorization}

Principal components analysis (PCA) and related techniques such as the singular value decomposition (SVD) are popular for the dimension reduction of a single data matrix $\bX: m \times n$, resulting in the low-rank approximation  $\bX \approx \bU\bV^T$. Here, $\bU$ are row loadings and $\bV$ are column scores that together explain variation in $\bX$.  There is also a growing literature on the simultaneous dimension reduction of multiple data matrices $\bX_{i}$ with size $m_{i}\times n$,  which estimate low-rank signals that are jointly shared across data matrices.   To capture joint variation, concatenated PCA assumes $\bX_i=\bU_i\bV^T$ for each matrix $\bX_i$, i.e., the scores are shared across matrices. The iCluster \citep{shen09} and irPCA \citep{liu16} approaches make this assumption for the integration of multi-source biomedical data.  Alternatively, more flexible approaches allow for structured variation that may be shared across matrices or specific to individual matrices. The  Joint and Individual Variations Explained (JIVE) method \citep{lock13} decomposes joint and individual low-rank signals across matrices via the decomposition $\bX_i=\bU_i\bV^T+\bW_i\bV_i^T+\bE_i$.   In the context of vertical integration, the joint and individual scores $\bV$ and $\bV_j$ have been applied to risk prediction \citep{kaplan17} and clustering \citep{hellton16} for high-dimensional data. Several related techniques { such as AJIVE \citep{feng18} and SLIDE \citep{gaynanova17}} have been proposed \citep{zhou16}, as well as extensions that allow the adjustment of covariates \citep{li17} or accommodate heterogeneity in the distributional assumptions for different sources \citep{li2018, zhu18}.


The aforementioned methods focus exclusively on data that share a single dimension (i.e., either horizontally or vertically), and extension to matrices that are linked both vertically and horizontally is not straightforward.  \citet{oconnell17} decompose shared and individual low-rank structure for three interlinked matrices $\bX,\bY,\bZ$ where $\bX$ and $\bY$ are shared vertically and $\bX$ and $\bZ$ are shared horizontally. However, their approach is not directly applicable to more general forms of bidimensionally linked data,  and it suffers from potential convergence to a local minimum of the objective during estimation.  

\subsection{Our Contribution}
We propose the first unified framework to decompose bidimensionally linked matrices into globally shared, horizontally shared (i.e., row-shared),  vertically shared (i.e., column-shared), or individual structural components. Our specific aims are to (i) separate shared and individual structures, (ii) separate the shared components into one of globally-shared, column-shared, or row-shared structures, and (iii) maintain the low-rank structures for the signals.  Our approach extends soft singular value thresholding (SSVT), i.e., nuclear norm penalization, for a single matrix. It requires optimizing a single convex objective function, which is relatively computationally efficient.  It also facilitates a simple and intuitive approach based on random matrix theory for model specification, rather than complex and computationally expensive procedures to select tuning parameters or model ranks.  Although our primary focus is bidimensional integration, our approach includes a novel method for vertical-only or horizontal-only integration as a special case.  We show in simulation studies that our method outperforms existing methods, including JIVE.

The rest of this article is organized as follows.  Section \ref{sec:meth} describes the proposed method, denoted by BIDIFAC (bidimensional integrative factorization) for bidimensionally linked matrices, and addresses estimation, tuning parameter selection and imputation algorithms. Section \ref{sec:sims} constructs simulated data under various scenarios and compares our method to existing methods in terms of structural reconstruction error and imputation performance. In Section \ref{sec:real}, we apply BIDIFAC to the breast cancer and NAT data obtained from TCGA and illustrate the utility of the model.  We conclude with some points of discussions in Section \ref{sec:conc}.

\section{Methods}
\label{sec:meth}

\subsection{Notations and Definitions}
{ Consider a set of $pq$ matrices $\lbrace \bA_{ij}: m_i \times n_j \mid  i=1,\dots, p, \; j=1,\dots,q\rbrace$, which may be concatenated to form the matrix  
\begin{align}
\bA_{00}= \left[\begin{array}{ccc} 
    \bA_{11} & \dots & \bA_{1q} \\
    \vdots & \ddots & \vdots \\
    \bA_{p1} & \dots      & \bA_{pq} 
    \end{array}\right], \label{arrange}
\end{align}
    where $\bA_{00}: m_0 \times n_0$ with with $m_0=\sum_{i=1}^p m_i$ and $n_0=\sum_{j=1}^q n_j$.  Analogously, we define the column-concatenated matrices $\bA_{i0}=[\bA_{i1}, \dots, \bA_{iq}]$ for $i=1,\ldots,p$ and the row-concatenated matrices $\bA_{0j}=[\bA_{1j}^T ,\dots, \bA_{pj}^T]^T$ for $j=1,\ldots,q$.}  
We first define terms to characterize the relationships among these data matrices.

\begin{definition}
{ In the arrangement \eqref{arrange}, a set of data matrices with the structure of  $\lbrace \bX_{ij}| i=1,\cdots, p, j=1,\cdots, q\rbrace$ follows a {\it $p\times q$ bidimensionally linked structure}}. The elements of $\lbrace \bX_{ij}|j=1,\cdots, q\rbrace$  are {\it row-shared} and the elements of  $\lbrace \bX_{ij}|i=1,\cdots, p\rbrace$ are {\it column-shared}. The elements of $\lbrace \bX_{ij}|i=1,\cdots,p, j=1\cdots,q\rbrace$  are {\it globally-shared} if every $\bX_{ij}$ in the set is column-shared with $\bX_{i^\prime j}$ and row-shared with $\bX_{ij^\prime}$ for $i^\prime=1,\cdots, p $ and $j^\prime=1,\cdots, q$.
\end{definition}

\subsection{Model Specification}
\begin{figure}[ht]
    \centering
    \includegraphics[scale=0.5]{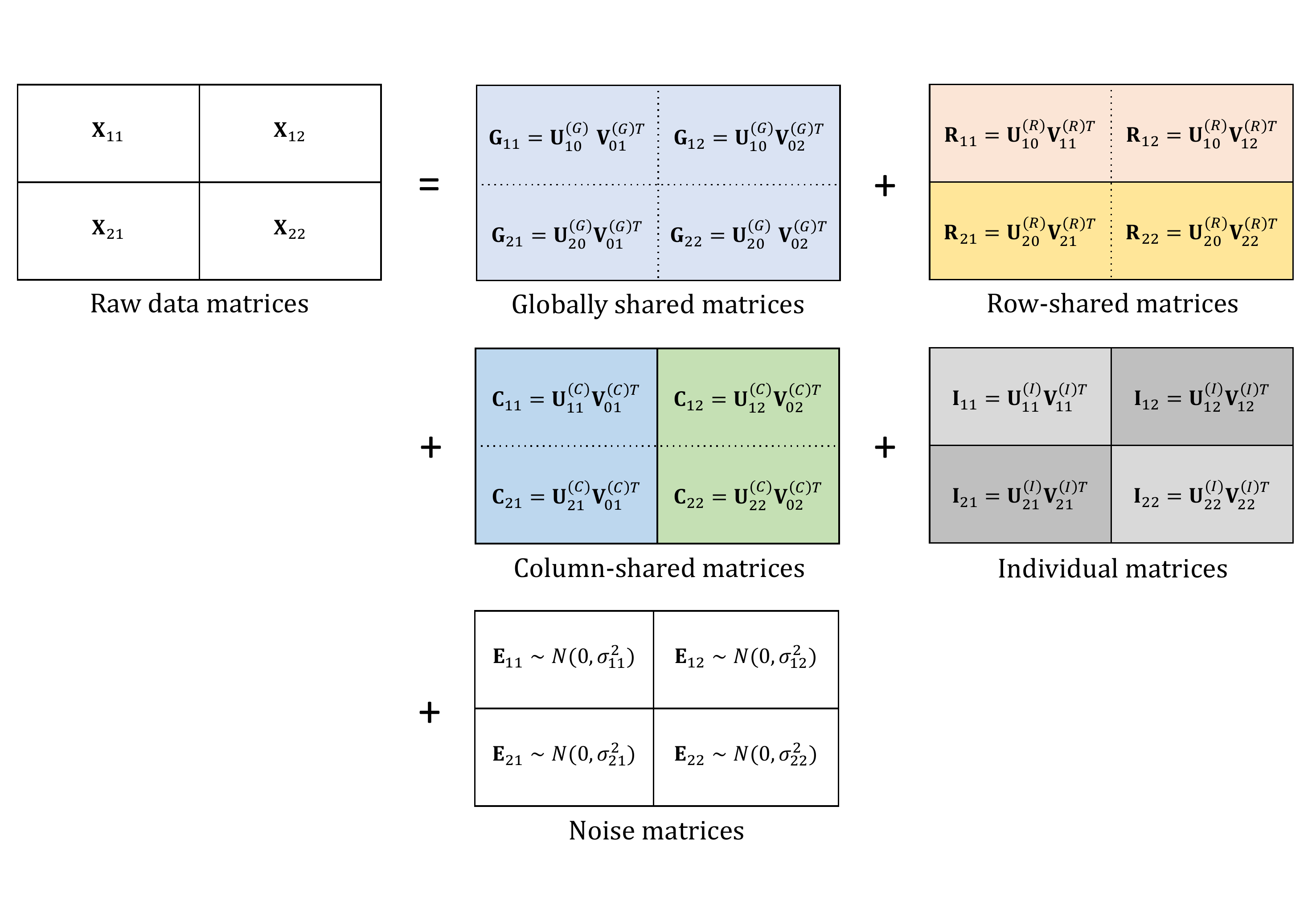}
    \caption{Overview of the proposed method, using the $2\times 2 $ linked structure.}
    \label{fig:1}
\end{figure}

We assume that each matrix is decomposed by $\bX_{ij}=\bS_{ij}+\bE_{ij}$, where  $\bS_{ij}$ is a low-rank signal matrix and { $\bE_{ij}$ is a full-rank white} noise.  We further assume that $\bS_{ij}$ can be decomposed as 
\begin{align}
\bS_{ij}=\bG_{ij}+\bR_{ij}+\bC_{ij}+\bI_{ij},
\end{align}
where $\bG_{ij}$ is globally shared structure, $\bR_{ij}$ is row-shared structure, $\bC_{ij}$ is column-shared structure, and $\bI_{ij}$ is individual structure for matrix $\bX_{ij}$.  The shared nature of the terms are apparent from their factorized forms. { Defining the parameter set as $\bThet=\lbrace \bG_{00}, \bR_{i0}, \bC_{0j}, \bI_{ij}|i=1,\cdots, p, j=1,\cdots,q \rbrace$, we write each term as a product of row \emph{loadings} $\bU$ and column \emph{scores} $\bV$:
\begin{align} \bG_{00} = \bU_{00}^{(G)} \bV_{00}^{(G) T}, \; \;  \bR_{i0} = \bU_{i0}^{(R)} \bV_{i0}^{(R) T},  \; \; \bC_{0j} = \bU_{0j}^{(C)} \bV_{0j}^{(C) T}, \; \; \bI_{ij} = \bU_{ij}^{(I)} \bV_{ij}^{(I) T} . \label{facform} \end{align}}
The loadings and scores are shared across matrices for the global components $\bG$, i.e., row-shared matrices have common global loadings $\bU_{i0}^{(G)}$, and column-shared matrices have common global scores $\bV_{0j}^{(G)}$.  The row-shared structures $\bR$ have common loadings, and the column-shared structures $\bC$ have common scores. { The dimensions of $\bU$ and $\bV$ depend on the global shared rank $r_{00}$, column-shared ranks $r_{0j}$, row-shared ranks $r_{i0}$, and individual ranks $r_{ij}$: $\bU_{ij}^{(\cdot)}~:~m_i\times r_{ij}$ and $\bV_{ij}^{(\cdot)}~:~n_j\times r_{ij}$ for $i=0,\cdots,p$, and $j=0,\cdots q$.
}
A diagram with the proposed notation for $2 \times 2$ linked structure is shown in Figure \ref{fig:1}.  

To understand the factorized forms in (\ref{facform}), it is instructive to consider their interpretation under the motivating example of Section~\ref{example}. Say  $\bX_{11}$ gives gene (mRNA) expression for tumor samples, $\bX_{12}$ gives gene expression for NAT samples, $\bX_{21}$ gives miRNA expression for tumor samples, and $\bX_{22}$ gives miRNA expression for NAT samples.  Then, the loadings of the global structures $\bG_{00}$  give trans-omic signatures $\bU_{00}^{(G)}$ that explain substantial variability across both tumor and NAT samples with associated scores $\bV_{00}^{(G)}$.  The loadings of the column-shared structures $\bC_{01}$  include trans-omic signatures $\bU_{01}^{(C)}$ that explain substantial variability in the tumor samples with associated scores $\bV_{01}^{(C)}$ but not the NAT samples.  The loadings of the row-shared structures $\bR_{10}$ include gene signatures $\bU_{10}^{(R)}$ that explain substantial variability across both tumor and NAT samples, but are unrelated to miRNA.  The loadings of the individual structures $\bI$ include gene signatures unrelated to miRNA that explain variability in only the tumor samples, $\bU_{11}^{(I)}$.  

 Marginally, $\bG_{i0}+\bR_{i0}$ denotes the components shared by $\bX_{i0}$ and $\bG_{0j}+\bC_{0j}$ denotes the components shared by $\bX_{0j}$. In practice, we assume the errors $\bE_{ij}$ are Gaussian white noise with mean zero and variance $\sigma_{ij}^2$.  For vertical integration, $q=1$ and the $\bG_{ij}$ and $\bR_{ij}$ terms are redundant with $\bC_{ij}$ and $\bI_{ij}$, respectively. { Thus, we suppress these terms and the decomposition of the model reduces to the ``joint and individual'' structures as in  one-dimensional factorization methods, including JIVE}.  Note, however, that we do not require pairwise orthogonality constraints on $\bG_{ij}$, $\bR_{ij}$, $\bC_{ij}$, and $\bI_{ij}$, which will be discussed in the remaining sections. 

\subsection{Estimation}
{ Without any penalization, estimation of $\bThet$ would have an identifiability issue. }
We first consider minimizing the sum of squared errors over all matrices with { different levels of  matrix $L_2$ penalties on  the $\lbrace \bU_{ij}^{(\cdot)},\bV_{ij}^{(\cdot)}\rbrace$}. 
Our objective function is
{
\begin{align}
    \nonumber    
    &f_1(\lbrace \bU_{ij}^{(\cdot)},\bV_{ij}^{(\cdot) }~|~ i=0,\dots, p, j=0,\dots,q, (i,j)\neq (0,0)\rbrace)\\
\nonumber     =&\sum_{i=1}^p\sum_{j=1}^q ||\bX_{ij}-\bU_{i0}^{(G)}\bV_{0j}^{(G)T}-\bU_{i0}^{(R)}\bV_{ij}^{(R)T}-\bU_{ij}^{(C)}\bV_{0j}^{(C)T}-\bU_{ij}^{(I)}\bV_{ij}^{(I)T}||_F^2\\ 
\nonumber    +&\lambda_{00} (||\bU_{00}^{(G)}||_F^2+||\bV_{00}^{(G)}||_F^2)+ \sum_{i=1}^p \lambda_{i0}(||\bU_{i0}^{(R)}||_F^2+||\bV_{i0}^{(R)}||_F^2)\\
     +&\sum_{j=1}^q \lambda_{0j}(||\bU_{0j}^{(C)}||_F^2+||\bV_{0j}^{(C)}||_F^2)+\sum_{i=1}^p\sum_{j=1}^q\lambda_{ij}(||\bU_{ij}^{(I)}||_F^2+||\bV_{ij}^{(I)}||_F^2) \label{eq:als}
\end{align}
}
where $||\cdot ||_F$ denotes the Frobenious norm and each $\lambda_{ij}$ is a non-negative penalty factor. We upper bound the ranks by setting $r_{ij}= \min (m_i, n_j)$  for  $i=0,\hdots,p$ and $j=0,\hdots, q$; the actual ranks of the solution may be lower, as discussed below.  
The objective function \eqref{eq:als} is a convex function of each of $\bU_{ij}^{(\cdot)}$ and $\bV_{ij}^{(\cdot)}$, given all the others fixed. One may use an alternating least squares (ALS) with a matrix $L_2$ penalty to iteratively update each of $\lbrace \bU_{00}^{(G)}, \bV_{00}^{(G)}\rbrace$, $\lbrace \bU_{i0}^{(R)}, \bV_{i0}^{(R)}\rbrace$, $\lbrace \bU_{0j}^{(C)}, \bV_{0j}^{(C)}\rbrace$, $\lbrace \bU_{ij}^{(I)}, \bV_{ij}^{(I)}\rbrace$ until convergence.


Alternatively, we reformulate \eqref{eq:als} and motivate our model using nuclear norm penalties.  A matrix $\bA: m \times n$ with ordered singular values $\delta_1, \delta_2,\ldots$ has nuclear norm
$||\bA||_* = \sum_{i=1}^{\text{min}\{m,n\}} \delta_i.$
We first present a well-known result on the equivalence of nuclear norm penalization and matrix factorization for a single matrix in Proposition~\ref{prop:1}.
\begin{prop} \label{prop:1} \citep{mazumder10} For a matrix $\bX: m\times n$,
\begin{align}
    \underset{\bU: n\times r,\bV: m\times r}{\min} ||\bX-\bU\bV^T||_F^2+\lambda (||\bU||_F^2+||\bV||_F^2) = \underset{\bZ: r(\bZ)\leq r}{\min} ||\bX-\bZ||_F^2 +2\lambda ||\bZ||_* \label{nucpen}
\end{align}
where $r=min(m,n)$.  Moreover, $\widehat{\bZ}=\widehat{\bU}\widehat{\bV}^T$, where $\widehat{\bZ}$ solves the right-hand side of \eqref{nucpen} and $\{\widehat{\bU},  \widehat{\bV}\}$ solves the left-hand side of \eqref{nucpen}.
\end{prop}
Proposition \ref{prop:1} depends on {Lemma~\ref{lem:1}}, which is also shown in \citet{mazumder10}.
\begin{lemma} \label{lem:1} For a matrix $\bZ:m\times n$, $\underset{\substack{\bU\bV^T=\bZ\\ \bU: m\times \min(m,n)\\ \bV:n\times \min(m,n)}}{\min}(||\bU||_F^2+||\bV||_F^2) = 2||\bZ||_*.$
\end{lemma}

We extend the nuclear norm objective in~\eqref{nucpen} to our context as follows:
\begin{align}
\nonumber   & f_2(\bThet)=\frac{1}{2}\sum_{i=1}^p\sum_{j=1}^q ||\bX_{ij}-\bG_{ij}-\bR_{ij}-\bC_{ij}-\bI_{ij}||_F^2 \\
   + & \lambda_{00}||\bG_{00}||_*+\sum_{i=1}^p \lambda_{i0}||\bR_{i0}||_*+\sum_{j=1}^q \lambda_{0j}||\bC_{0j}||_*+\sum_{i=1}^p\sum_{j=1}^q \lambda_{ij}||\bI_{ij}||_*.\label{eq:ssvt}
\end{align}
Theorem~\ref{thm:1} establishes the equivalence of \eqref{eq:als} and \eqref{eq:ssvt}, with proof in Appendix~\ref{proofs}.
\begin{theorem}
\label{thm:1}
{
Let \[\widehat{\bThet}_1=\lbrace\widehat{\bU}_{i0}^{(G)}\widehat{\bV}_{0j}^{(G)T}, \widehat{\bU}_{i0}^{(R)}\widehat{\bV}_{ij}^{(R)T}, \widehat{\bU}_{ij}^{(C)}\widehat{\bV}_{0j}^{(\bC)T}, \widehat{\bU}_{ij}^{(I)}\widehat{\bV}_{ij}^{(I)T} \mid i=1\ldots p, j=1,\ldots q\rbrace\] minimize \eqref{eq:als}.  Then, \[\widehat{\bThet}_2 = \lbrace \widehat{\bG}_{ij}, \widehat{\bR}_{ij}, \widehat{\bC}_{ij}, \widehat{\bI}_{ij} \mid i=1\ldots p, j=1,\ldots q \rbrace\] minimizes \eqref{eq:ssvt}, where  $\widehat{\bG}_{ij}=\widehat{\bU}_{i0}^{(G)}\widehat{\bV}_{0j}^{(G)T}, \widehat{\bR}_{ij}=\widehat{\bU}_{i0}^{(R)}\widehat{\bV}_{ij}^{(R)T}, \widehat{\bC}_{ij}=\widehat{\bU}_{ij}^{(C)}\widehat{\bV}_{0j}^{(C)T},$ and  $\widehat{\bI}_{ij}=\widehat{\bU}_{ij}^{(I)}\widehat{\bV}_{ij}^{(I)T}$. }
\end{theorem}

The objective \eqref{eq:ssvt} has several advantages.  First, the function is convex, which we state in Theorem~\ref{thm:2} and prove in Appendix~\ref{proofs}.
\begin{theorem}
The objective $f_2(\cdot)$ in~\eqref{eq:ssvt} is convex over its domain.  
\label{thm:2}
\end{theorem}
Fortunately,  minimizing one term ($\bG_{00}, \bR_{i0}, \bC_{0j}$, or $\bI_{ij}$) with the others fixed is straightforward via soft singular value thresholding (SSVT).  We state the well-known equivalence between nuclear norm penalization and SSVT in Proposition~\ref{prop:2}; for a proof see~\citet{mazumder10}.
\begin{prop} \label{prop:2} If the SVD of $\bX$ is $\bU_X\bD_X\bV_X^T$ and $\bD_X$ has diagonal entries $\delta_1\geq \dots \geq \delta_r\geq 0$, the solution for $\bZ$ in~\eqref{nucpen} is equal to $\widehat{\bZ}=\bU_X \bD_X(\lambda)\bV_{X}^T$, where $\bD_X(\lambda)$ is a diagonal matrix with $\delta_1, \dots, \delta_r$ replaced by $\max(\delta_1-\lambda,0), \ldots, \max (\delta_r-\lambda, 0)$, respectively.
\end{prop}
We use an \emph{iterative soft singular value thresholding} (ISSVT) algorithm to solve~\eqref{eq:ssvt}, applying Proposition~\ref{prop:2} to the appropriate residual matrices for $\bG_{00}, \bR_{i0}, \bC_{0j}$, or $\bI_{ij}$.   For example, $\widehat{\bI}_{ij}$ and $\widehat{\bR}_{i0}$ are obtained by soft-thresholding the singular values of $\bX_{ij}-\widehat{\bG}_{ij}-\widehat{\bR}_{ij}-\widehat{\bC}_{ij}$ and $\bX_{i0}-\widehat{\bG}_{i0}-\widehat{\bI}_{i0}$ towards 0, respectively. {This iterative algorithm is guaranteed to converge to a coordinatewise-minimum, and convexity implies that it will be a global minimum if it is a local minimum.  In practice, we find that iterative algorithms to solve either \eqref{eq:als} or \eqref{eq:ssvt} converge to the same solution and are robust to their initial values.}
The detailed algorithm of ISSVT is provided in Appendix~\ref{algorithm}. 
The resulting global, column-shared, row-shared and individual terms of the decomposition will have reduced rank depending on the penalty factors $\lambda_{ij}$.  Moreover, this relation between the penalty factors and the singular values motivates a straightforward choice of tuning parameters using random matrix theory described in Section \ref{sec:tuningparameters}.

{ 
ISSVT can also be represented as a blockwise coordinate descent algorithm for the ALS objective \eqref{eq:als} (see Appendix~\ref{algorithm}), 
and it converges faster than ALS, so we use it as our default algorithm.  However,}
the ALS approach may be extended to certain related contexts.  For example, it can incorporate sparsity in the loadings via an additional penalty (e.g., $L_1$) on $\bU_{ij}^{(\cdot)}$. Also, with only three linked matrices $\bX_{11},\bX_{12}$ and $\bX_{21}$, as in \citet{oconnell17}, \eqref{eq:ssvt} cannot properly construct the globally shared components.  The formulation of \eqref{eq:als} with an ALS algorithm can handle such contexts.

\subsection{Data Pre-Processing}
\label{sec:preprocessing}
In practice, the data matrices $\bX_{ij}$ may have very different levels of variability or be measured on different scales.  Thus, a straightforward application of the objective \eqref{eq:als} or \eqref{eq:ssvt} is not appropriate without further processing.    By default we center the matrices to have mean $0$, so that each matrix has the same baseline.  To resolve issues of scale, we propose dividing each data matrix $\bX_{ij}, i,j>0$ by an estimate of the square root of its noise variance $\hat{\sigma}_{ij}$, denoted by $\bX_{ij, scale}$.  We discuss estimating the Gaussian noise variance in Section \ref{sec:tuningparameters}.   After scaling, each matrix has homogeneous unit noise variance, which motivates the proposed penalties. After all components, denoted by $\bG_{ij,scale}, \bR_{ij,scale}, \bC_{ij,scale},$ and $\bI_{ij,scale}$, are estimated, we transform the results back to the original scale by multiplying each matrix by $\hat{\sigma}_{ij}$. 

We comment on estimating the noise variance $\sigma_{ij}^2$. Without any signal or with a very weak signal, the standard deviation of $\mbox{vec}(\bX_{ij})$, denoted as $\hat{\sigma}_{ij}^{SD}$, provides a nearly unbiased estimate of $\sigma_{ij}$. However, this estimate is biased and overly conservative with a high signal-to-noise ratio. An alternative is to use random matrix theory, and estimate $\sigma_{ij}$ by minimizing the Kolmogorov-Smirnov distance between the theoretical and empirical  distribution functions of singular values, as in \citet{shabalin13}. Their estimate $\hat{\sigma}_{ij}^{KS}$ is based on grid-search on a candidate set of $\sigma$. Recently, \citet{garvish17} proposed another estimator $\hat{\sigma}_{ij}^{MAD}$, also based on random matrix theory, which is defined as the  median of the singular values of $\bX_{ij}$ divided by the square root of the median of the Marcenko-Pastur distribution. Our simulations, not shown here, revealed that both $\hat{\sigma}_{ij}^{KS}$ and $\hat{\sigma}_{ij}^{MAD}$ well approximate  the standard deviation of a true noise matrix when the data matrix consists of low rank signal, and we use $\hat{\sigma}_{ij}^{MAD}$ as a default throughout this paper for its simplicity. {From here, we assume that BIDIFAC is applied to the data matrices with  $\sigma_{ij}^2=1$}.

\subsection{Summarizing Results}
\label{sec:preprocessing}
Given that the mean of each matrix is 0, we propose proportion of variance explained $(R_{\bX_{ij}}^2(\cdot))$ as a summary statistic. For example, 
\begin{align}
    R_{\bX_{ij}}^2(\bG_{ij})=1-\dfrac{||\bX_{ij}-\bG_{ij}||_F^2}{||\bX_{ij}||_F^2} \label{eq:0}
\end{align}
provides a measure of the proportion of variability explained by the globally shared component.  However, because orthogonality is not explicitly enforced in BIDIFAC,  this equality does not hold in general and $R^2(\cdot)$ is not necessarily additive across terms (e.g., $R_{\bX_{ij}}^2(\bG_{ij}+\bC_{ij})\neq R_{\bX_{ij}}^2(\bG_{ij})+R_{\bX_{ij}}^2(\bC_{ij})$).

\subsection{Selecting Tuning Parameters}
\label{sec:tuningparameters}
The performance of the proposed method depends heavily on the choice of tuning parameters. In the literature, there are several approaches to select ranks in the context of vertical integration, including permutation testing \citep{lock13}, BIC \citep{oconnell16}, and cross-validation \citep{li17}.  In our context, the issue of rank selection is analogous to selecting the tuning parameters $\lambda_{ij}$.  Although  cross-validation is a natural way of selecting tuning parameters in penalized regression, our objective involves too many parameters ($1+p+q+pq$) to be computationally feasible. Moreover, despite the rich literature on cross-validating SVD for a single matrix, it is not clear how to define the training and test set (e.g., randomly select cells, rows, columns, a whole matrix, etc). A general description of the difficulties in cross-validating matrices is provided by \citet{owen09}.

{ Admitting that cross validation in BIDIFAC is not straightforward and inefficient,  we provide an alternative approach to select the tuning parameters based on random matrix theory. We first construct necessary conditions for each element of $\widehat{\bThet}$ to be nonzero.
\begin{prop}
The following conditions are necessary to allow for non-zero $\widehat{\bG}_{00}$, $\widehat{\bR}_{i0}$, $\widehat{\bC}_{0j}$, and $\widehat{\bI}_{ij}$: 
   \begin{enumerate}
   \item $ \mbox{max}_j \lambda_{ij} < \lambda_{i0} < \sum_j  \lambda_{ij}$ for $i=1,\hdots,p$ and $\mbox{max}_i \lambda_{ij} < \lambda_{0j} < \sum_i  \lambda_{ij} $ for $j=1,\hdots,q$
   \item $\mbox{max}_i \lambda_{i0} < \lambda_{00} < \sum_i  \lambda_{i0} $
   and $\mbox{max}_j \lambda_{0j} < \lambda_{00} < \sum_j \lambda_{0j} $.
   \end{enumerate} \label{prop3}
\end{prop}
We provide a proof of Proposition \ref{prop3} in Appendix~\ref{proofs}.} Without loss of generality, suppose that each cell of $\bX_{ij}$ has Gaussian noise with unit variance ($\sigma_{ij}^2=1$), which is independent within each matrix and across matrices. Based on random matrix theory, we propose using the following penalty factors:
\begin{align}
    \lambda_{ij}=\sqrt{m_i}+\sqrt{n_j},~~~\mbox{where}~~~i=0,\dots, p, ~~\mbox{and}~~~j=0,\dots,q. \label{eq:2}
\end{align} 
{ It is straightforward to show that our choice of tuning parameters meets the necessary requirement.  Also, }under the aforementioned assumptions, $\sqrt{m_i}+\sqrt{n_j}$ provides a tight upper bound for the largest singular value of $\bE_{ij}$ \citep{rudelson10}.  Thus, without any shared structure, the motivation for \eqref{eq:2} is apparent by considering the penalty as a soft-thresholding operator on the singular values in Proposition \ref{prop:2}. The penalty \eqref{eq:2} is also used in the single matrix reconstruction method in \citet{shabalin13}. The penalties for the shared components are decided analogously because the stack of column-/row- shared matrices are also Gaussian random matrices with unit noise variance. For example, the penalty for $\bR_{i0}$ is determined given by an estimate of the largest singular value of its concatenated noise matrix $\bE_{i0}$ with unit variance: $\sqrt{m_i}+\sqrt{n_0}$. 

\subsection{Imputation}
\label{sec:imputation}
A convenient feature of BIDIFAC is its potential for missing value imputation. For single block data, PCA or related low-rank factorizations can be used to impute missing values by iteratively updating missing entries with their low-rank approximation, and this approach has proven to be very accurate in many applications \citep{kurucz07}. An analogous algorithm has been used for imputation with joint matrix factorization \citep{oconnell17}, and this approach readily extends to BIDIFAC.  Importantly, this allows for the imputation of data that are missing an entire row or column within a block, via an expectation-maximization (EM) approach. Our imputation algorithm is presented below.

\begin{enumerate}
    \item Let $\mathcal{I}_{ij}=\lbrace (r,s) ~|~ \bX_{ij}[r,s] \mbox{ is missing}\rbrace$. For each matrix, initialize the missing values by the column- and/or row-wise mean and denote the initial matrix by $\widehat{\bX}_{ij}^{(old)}$.

    \item  
    \begin{enumerate}
        \item Maximization: Apply BIDIFAC to $\lbrace \widehat{\bX}_{ij}^{(old)}\rbrace_{i,j=1}^{p,q}$.
        \item Expectation: For $(r,s)\in\mathcal{I}_{ij}$,  replace $\widehat{\bX}_{ij}^{(old)}[r,s]$ with $\widehat{\bS}_{ij}[r,s]$ and denote the imputed matrix by $\widehat{\bX}_{ij}^{(new)}$. 
    \end{enumerate}
        \item If $\sum_{i,j=1}^{p,q}\sum_{(r,s)\in\mathcal{I}_{ij}}|\widehat{\bX}_{ij}^{(old)}[r,s]-\widehat{\bX}_{ij}^{(new)}[r,s]|^2<\epsilon$, the algorithm converges. If not, re-apply (2) after replacing $\widehat{\bX}_{ij}^{(old)}$ by $\widehat{\bX}_{ij}^{(new)}$.    
\end{enumerate}
To improve computational efficiency, each maximization step uses the   $\widehat{\bThet}$ from the previous maximization step as starting values.  {The imputation scheme can be used to impute entries or entire rows or columns of the constituent data matrices; however, in all cases the missing entities (entries, rows or columns) must be missing at random.}

Our algorithm can be considered a regularized EM algorithm, using a model-based motivation for the objective function~\eqref{eq:als}.
Because each component is estimated as a product of two matrices, it is naturally translated as a probabilistic matrix factorization.  The unpenalized objective with $\lambda_{ij}=0$ for all $i,j$ maximizes a Gaussian likelihood model if the noise variances are the same across the matrices (i.e., $\sigma^2=\sigma_{11}^2=\dots=\sigma_{pq}^{2}$).   With penalization, the approach is analogous to maximizing the posterior distribution in a Bayesian context with Gaussian prior on the terms of  $\bU_{ij}^{(\cdot)}$ and $\bV_{ij}^{(\cdot)}$.  Specifically, if the entries of $\bU_{ij}^{(\cdot)}$ and $\bV_{ij}^{(\cdot)}$ are independent Gaussian with mean $0$ and variance $\sigma^2/\lambda_{ij}$, then minimizing~\eqref{eq:als} is analogous to finding the posterior mode \citep{mnih08}.  This provides a theoretical foundation for the iterative imputation algorithm based on expectation-maximization (EM) algorithm. It is closely related to softImpute, proposed by \citet{mazumder10}, a computationally efficient imputation algorithm for a single matrix based on Proposition \ref{prop:1}.  {Prediction intervals for the imputed values may be obtained via a resampling approach, as described in Web Appendix E of \citet{oconnell17}.}

\section{Simulation Studies}
\label{sec:sims}

\subsection{Simulation Setup}
In this section, we compare our model to existing factorization methods using simulated data. Because competing approaches apply only to  uni-dimensionally linked matrices (vertical or horizontal), we constructed two simulation designs with $p=q=2$ for proper comparison. Design 1 does not include any row-shared or globally shared structure and may be considered as two separate sets of vertically-linked  matrices. Design 2 includes all linked structures that are represented in our model: global, column-shared, row-shared and individual.

For each simulation design, we used existing methods to compare performance. First, we fit 2 separate JIVE models to $\lbrace \bX_{11}, \bX_{21}\rbrace$ and $\lbrace \bX_{12}, \bX_{22}\rbrace$, using both (i) true marginal ranks for joint and individual components and (ii) rank selection based on permutation testing, denoted by JIVE(T) and JIVE(P) respectively. The ``true marginal rank'' in this context means $r(\bG_{ij}+\bC_{ij})$ for the joint component and $r(\bR_{ij}+\bI_{ij})$ for the individual components given that $j$ is fixed. { We similarly apply the AJIVE and SLIDE  methods, where we used the rank of $\bS_{ij}$ as the initial rank for AJIVE.} We also consider an approach that is analogous to BIDIFAC but reduced to vertical integration only, i.e., with $\widehat{\bG}_{ij}$ and $\widehat{\bR}_{ij}$ set to ${\bf 0}_{m_i\times n_j}$, denoted as UNIFAC in this paper.  We fit UNIFAC to $\lbrace \bX_{11}, \bX_{21}\rbrace$ and $\lbrace \bX_{12}, \bX_{22}\rbrace$ separately. For notational simplicity, we denote the joint and individual components estimated by  JIVE(P), JIVE(T), AJIVE, SLIDE and UNIFAC by $\widehat{\bC}_{ij}$ and $\widehat{\bI}_{ij}$. 
We also applied soft singular value thresholding for each single matrix with the corresponding imputation algorithm, softImpute, with tuning parameters decided as in Section~\ref{sec:tuningparameters} $(\lambda_{ij}=\hat{\sigma}_{ij}\cdot (\sqrt{m_i}+\sqrt{n_j}))$. We denote this approach by SVD(soft). We also consider the performance of the hard-thresholding low-rank approximation of each matrix, denoted by SVD(T), using the true marginal rank for a single matrix ($r(\bS_{ij})$). {Similar to the above, SVD(soft) and SVD(T) estimate $\bI_{ij}$ (or, equivalently, $\bS_{ij}$) components only and assume $\widehat{\bG}_{ij}=\widehat{\bR}_{ij}=\widehat{\bC}_{ij}={\bf 0}_{m_i\times n_j}$. }BIDIFAC, UNIFAC, and SVD(soft) are soft-thresholding methods, while JIVE(T), JIVE(P), and SVD(T) are based on hard-thresholding.

In our simulation studies, the number of rows and columns for each matrix was set to $100$: $m_i=n_j=100$. The rank of the total signal in each matrix, $\bS_{ij}=\bG_{ij}+\bR_{ij}+\bC_{ij}+\bI_{ij}$, was $10$. This total rank was distributed across each of the $4$ terms (or the two terms $\bC_{ij}$ and $\bI_{ij}$ for Design 1) via a multinomial distribution with equal probabilities. For clarity of the simulation studies and to allow for comparison with other methods, we enforced orthogonality among the shared structures, both within each matrix (i.e., $\bG_{11}$ and $\bC_{11}$ are orthogonal) and across matrices (i.e., $\bG_{11}$ and $\bG_{12}$ are orthogonal). Note, however, that our model does not enforce orthogonality when estimating parameters.

Each signal matrix, $\bS_{ij}$, was generated by applying SVD to a Gaussian random matrix with mean 0 and unit variance,  denoted as $\bY=\bU_Y\bD_Y\bV_Y^T$. Then $\bU_Y$ and $\bV_Y$ were rearranged accordingly to guarantee orthogonality and a $2\times 2$ linked structure.  The singular values were randomly permuted within each shared component to allow for heterogeneity in the size of the joint signal across matrices, e.g.,  $\bC_{11}$ and $\bC_{21}$ have the same loadings and scores but with different order. 
Each signal matrix $\bS_{ij}$ was standardized to have $||\bS_{ij}||_F=1$.  Finally, independent Gaussian noise was added to each signal matrix, where the noise variance was decided by the signal-to-noise ratio (SNR) defined by $1/(\sigma_{ij}\cdot\sqrt{m_i\cdot n_j})$. We first considered three SNRs in our simulation studies: 0.5, 1, and 2. For the true structure $\bS_{ij}$ the expected value of $R_{\bX_{ij}}^2(\bS_{ij})$ is $\mbox{SNR}^2/(1+\mbox{SNR}^2)$, when SNRs are the same for all matrices. We also considered a scenario where the SNR is randomly selected separately for the different matrices, uniformly between 0.5 and 2, to accommodate heterogeneous noise variances.

We compared the performance of our method and the competing methods from two perspectives: prediction error and imputation performance. We computed prediction error for each term in the decomposition ($\bG, \bR, \bC$ or $\bI$) as the relative reconstruction error:
\begin{align}
    \mbox{PredErr}(\widehat{\bG})=\dfrac{\sum_{i=1,j=1}^{p,q} ||\bG_{ij}-\widehat{\bG}_{ij}||_F^2}{\sum_{i=1,j=1}^{p,q}  ||\bG_{ij}||_F^2}.\label{eq:prederr}
\end{align}
For a fair comparison between our method and the existing methods for uni-dimensionally shared matrices, we also report $\mbox{PredErr}(\widehat{\bG}+\widehat{\bC})$, $\mbox{PredErr}(\widehat{\bR}+\widehat{\bI})$. Finally, we also report $\mbox{PredErr}(\widehat{\bS})$ to evaluate the overall signal reconstruction performances. 

For imputation, we considered three scenarios where in each matrix (i) 200 randomly selected cells are missing, (ii) 2 columns are missing, and (iii) 2 rows are missing. To foster borrowing information from the shared structures in (ii) and (iii), there was no overlapping row/column that are missing simultaneously in shared matrices. { To reduce computation cost, imputation using JIVE(P) used fixed ranks determined by the complete data, resulting in  slightly inflated imputation performances}. We evaluated the imputation performance using the scaled reconstruction error for missing cells, defined by
\begin{align}
    \mbox{ImputeErr}=\dfrac{\sum_{i=1,j=1}^{p,q}\sum_{(r,s)\in \mathcal{I}_{ij}} |\bS_{ij}[r,s]-\widehat{\bS}_{ij}[r,s]|^2}{\sum_{i=1,j=1}^{p,q}\sum_{(r,s)\in \mathcal{I}_{ij}} |\bS_{ij}[r,s]|^2}. \label{eq:imperr}
\end{align}

\subsection{Results}
We repeated each simulation 200 times and averaged the performance. The results are shown in Tables \ref{tbl1} and \ref{tbl2}. 
We summarize the results below from a few perspectives.

\begin{table}[!ht]
\centering
\caption{Summary of the simulation studies for Design 1, where prediction and imputation errors are computed using Equations (\ref{eq:prederr}) and (\ref{eq:imperr}) respectively. }\label{tbl1}
\begin{tabular}{clcccccccccc}
  \hline
  &  &  \multicolumn{7}{l}{Prediction} & \multicolumn{3}{l}{Imputation}\\ \cline{3-12}
SNR &Model &  $G$ & $R$ & $C$ & $I$ & $G+C$ & $R+I$ & $S$ & Cell & Column & Row \\ 
  \hline
0.5 &BIDIFAC & $-$ & $-$ & 0.81 & 0.87 & 0.80 & 0.87 &  0.81 & 0.86 & 0.95 & 1.00\\  
&UNIFAC & $-$ & $-$ &   0.80 & 0.87 & 0.80 & 0.87 & 0.81  & 0.86 & 0.94 & 1.00 \\  
&JIVE(T) & $-$ & $-$ & 1.01 & 1.40 & 1.01 & 1.40 & 1.05 & 1.49 & 1.13 & 1.02 \\  
&JIVE(P) & $-$ & $-$ & 1.09 & 2.67 & 1.09 & 2.67 & 1.33 & 2.05 & 1.13 & 1.02 \\
 &AJIVE(T) & $-$ & $-$ & 4.58 & 1.00 & 4.58 & 1.00 & 1.48 & $-$ & $-$ & $-$ \\
&SLIDE & $-$ & $-$ & 0.97 & 1.02 & 0.97 & 1.02 & 0.84 & $-$ & $-$ & $-$ \\
&SVD(T) & $-$ & $-$ & 1.00 & 3.61 & 1.00 & 3.61 & 1.20 & 1.87 & 1.02 & 1.02\\ 
&SVD(soft) & $-$ & $-$ & 1.00 & 0.89 & 1.00 & 0.89 & 0.85& 0.90 & 1.00 & 1.00\\   \hline
1 & BIDIFAC & $-$ & $-$ & 0.35 & 0.41 & 0.35 & 0.41 & 0.36 & 0.43 & 0.80 & 1.00\\ 
 &UNIFAC & $-$ & $-$ & 0.35 &0.41 &0.35 &0.41 &0.36  & 0.43 & 0.79 & 1.00 \\  
 &  JIVE(T) & $-$ & $-$ & 0.18 & 0.23 & 0.18 & 0.23 & 0.19 & 0.25 & 0.63 & 1.01 \\ 
 &  JIVE(P) & $-$ & $-$ & 0.47 & 0.54 & 0.47 & 0.54 & 0.23 & 0.40 & 0.78 & 1.01 \\
  &AJIVE(T) & $-$ & $-$ & 1.98 & 1.00 & 1.98 & 1.00 & 0.30 & $-$ & $-$ & $-$ \\
 &SLIDE & $-$ & $-$ & 0.19 & 0.22 & 0.19 & 0.22 & 0.19 & $-$ & $-$ & $-$ \\
 &SVD(T) & $-$ & $-$ &1.00 & 1.73 & 1.00 & 1.73 & 0.22 & 0.30 & 1.01 & 1.01\\ 
&SVD(soft) & $-$ & $-$ & 1.00 & 0.66 & 1.00 & 0.66 & 0.40 & 0.49 & 1.00 & 1.00\\ \hline
2 &  BIDIFAC & $-$ & $-$ & 0.11 & 0.13 & 0.11 & 0.13 & 0.11 & 0.16 & 0.65 & 1.00 \\
 &UNIFAC & $-$ & $-$ &  0.11 &0.13& 0.11& 0.13& 0.11  & 0.16 & 0.65  & 1.00 \\  
 &  JIVE(T) & $-$ & $-$ & 0.04 & 0.05 & 0.04 & 0.05 & 0.04 & 0.05 & 0.54 & 1.01 \\ 
 &  JIVE(P) & $-$ & $-$ & 0.35 & 0.32 & 0.35 & 0.32 & 0.05 & 0.14 &  0.81 & 1.01 \\ 
 &AJIVE(T) & $-$ & $-$ & 0.07 & 0.08 & 0.07 & 0.08 & 0.04 & $-$ & $-$ & $-$ \\
 &SLIDE & $-$ & $-$ & 0.05 & 0.06 & 0.05 & 0.06 & 0.06 & $-$ & $-$ & $-$ \\
  &SVD(T) & $-$ & $-$ & 1.00 & 1.37 & 1.00 & 1.37 & 0.05 & 0.06 & 1.01 & 1.01 \\   
&SVD(soft) & $-$ & $-$ &1.00 & 0.74 & 1.00 & 0.74 & 0.13 & 0.18 & 1.00 & 1.00\\ 
 \hline
Mixed & BIDIFAC & $-$ & $-$ &  0.37 & 0.41 & 0.37 & 0.41 & 0.31&  0.38 & 0.80 & 1.00\\ 
  &UNIFAC & $-$ & $-$ &  0.37 &0.41& 0.37& 0.41 & 0.31  & 0.36 & 0.77  & 1.00 \\  
 & JIVE(T) & $-$ & $-$ & 0.20 & 0.27 & 0.20 & 0.27 & 0.22& 0.29 & 0.64 & 1.01 \\ 
 & JIVE(P) & $-$ & $-$ & 0.55 & 0.73 & 0.55 & 0.73 & 0.29 & 0.47 & 0.83 & 1.01\\ 
 & AJIVE(T) & $-$ & $-$ & 2.28 & 0.92 & 2.28 & 0.92 & 0.64 & $-$ & $-$ & $-$ \\
 &SLIDE & $-$ & $-$ & 0.51 & 0.53 & 0.51 & 0.53 & 0.38 & $-$ & $-$ & $-$ \\
 & SVD(T) & $-$ & $-$ & 1.00 & 1.78 & 1.00 & 1.78 & 0.74 & 0.35 & 1.01 & 1.01 \\ 
 & SVD(soft) & $-$ & $-$ &  1.00 & 0.70 & 1.00 & 0.70 & 0.59 & 0.43 & 1.00 & 1.00\\ 
\hline
\end{tabular}
\end{table}

\begin{table}[!ht]
\centering
\caption{Summary of the simulation studies for Design 2.}\label{tbl2}
\begin{tabular}{clcccccccccc}
  \hline
  &  &  \multicolumn{7}{l}{Prediction} & \multicolumn{3}{l}{Imputation}\\ \cline{3-12}
SNR &Model  & $G$ & $R$ & $C$ & $I$ & $G+C$ & $R+I$ & $S$ & Cell & Column & Row \\ 
  \hline
0.5 & BIDIFAC  & 0.67 & 0.60 & 0.81 & 0.89 & 0.73 & 0.83 & 0.76 &  0.81 & 0.91 & 0.91\\
&UNIFAC &  1.00 & 1.00 &  0.85 &0.91& 0.80& 0.87& 0.82  & 0.86 & 0.95  & 1.00 \\  
  &JIVE(T)  & 1.00 & 1.00 & 3.05 & 3.86 & 1.00 & 1.38 & 1.04 & 1.47 & 1.14 & 1.02  \\ 
  &JIVE(P)  & 1.00 & 1.00 & 2.60 & 6.58 & 1.12 & 2.51 & 1.36 & 2.13 & 1.17 & 1.02 \\  
  &AJIVE(T) &  1.00  & 1.00 & 10.74 &  1.00 &  4.05 &  1.00 &  1.47 & $-$& $-$& $-$\\
  &SLIDE &  1.00& 1.00& 1.00& 1.27& 0.96 & 0.97 & 0.85& $-$& $-$& $-$\\
  &SVD(T) & 1.00 &1.00  & 1.00 & 8.51 & 1.00 & 3.32 & 1.19  & 1.86 & 1.02 & 1.02\\ 
  &SVD(soft) & 1.00 & 1.00 & 1.00 & 0.95 & 1.00 & 0.89 & 0.86 & 0.90 & 1.00 & 1.00 \\
  \hline
1 &  BIDIFAC & 0.26 & 0.26 & 0.34 & 0.42 & 0.30 & 0.37 & 0.32 &  0.39 & 0.77 & 0.77 \\
 &UNIFAC &  1.00 & 1.00  & 0.65& 0.68& 0.34& 0.41 &0.36 & 0.44 & 0.79  & 1.00 \\  
 &  JIVE(T) & 1.00 & 1.00 & 1.62 & 1.76 & 0.18 & 0.23 & 0.19 & 0.25 & 0.63 & 1.01 \\
 &  JIVE(P) & 1.00 & 1.00 & 1.72 & 2.18 & 0.45 & 0.48 & 0.23 & 0.40 & 0.77 & 1.01 \\
 & AJIVE(T) & 1.00 & 1.00 & 5.77 & 1.00 & 1.76 & 1.00 & 0.30 & $-$& $-$& $-$\\
  &SLIDE & 1.00 & 1.00 & 0.99 & 0.93 & 0.19 & 0.22 & 0.19  & $-$& $-$& $-$\\
 & SVD(T) & 1.00  & 1.00 & 1.00 & 4.82 & 1.00 & 1.56 & 0.22 & 0.29 & 1.01 & 1.01 \\ 
 & SVD(soft) & 1.00 & 1.00 & 1.00 & 1.19 & 1.00 & 0.63 & 0.40 & 0.49 & 1.00 & 1.00 \\ 
\hline
2 &  BIDIFAC & 0.09 & 0.09 & 0.11 & 0.14 & 0.10 & 0.12 & 0.10 &  0.14 & 0.63 & 0.64\\ 
  &UNIFAC &  1.00 & 1.00  & 0.76 &0.76 &0.11 &0.13 &0.11   &0.16 & 0.65  & 1.00 \\  
 &  JIVE(T) & 1.00 & 1.00 & 1.33 & 1.39 & 0.04 & 0.05 & 0.04 & 0.06 & 0.54 & 1.01 \\
 &  JIVE(P) & 1.00 & 1.00 & 1.42 & 1.70 & 0.29 & 0.25 & 0.05 & 0.14 &  0.78 &  1.01  \\ 
 &  AJIVE(T) & 1.00& 1.00& 1.39& 1.57 & 0.05 & 0.06& 0.04 &$-$ & $-$ & $-$\\
 &SLIDE & 1.00 & 1.00 & 1.02& 1.19& 0.05& 0.06& 0.06&  $-$ & $-$ & $-$\\
 &SVD(T) & 1.00 & 1.00 & 1.00 & 4.07 & 1.00 & 1.22 & 0.05 & 0.06 & 1.01 &  1.01\\ 
 & SVD(soft) & 1.00 & 1.00 & 1.00 & 2.00 & 1.00 & 0.67 & 0.13 & 0.18 & 1.00 &1.00 \\ 
 \hline
Mixed&BIDIFAC & 0.30 & 0.28 & 0.37 & 0.48 & 0.31 & 0.35& 0.27  & 0.33 & 0.77 & 0.78  \\ 
  &UNIFAC &  1.00 & 1.00 &  0.69 &0.79& 0.36 &0.40 &0.31   & 0.37 & 0.77  & 1.00 \\
  &JIVE(T) & 1.00 & 1.00 & 1.65 & 1.83 & 0.20 & 0.27 & 0.22 & 0.29 & 0.64 & 1.01 \\ 
  &JIVE(P) & 1.00 & 1.00 & 1.85 & 2.53 & 0.53 & 0.65 & 0.30 & 0.46 & 0.87 & 1.01\\
  &AJIVE(T) & 1.00 & 1.00 & 6.35 & 1.11 & 2.07 & 0.90 & 0.60 & $-$ & $-$ & $-$\\
  &SLIDE & 1.00 & 1.00 & 1.01 & 1.37 & 0.49 & 0.49 & 0.34 & $-$&$-$&$-$\\
  &SVD(T) & 1.00 & 1.00 & 1.00 & 4.94 & 1.00 & 1.61 & 0.25 & 0.36 & 1.01 & 1.01 \\
  & SVD(soft) & 1.00 & 1.00 & 1.00 & 1.39 & 1.00 & 0.66 & 0.36 & 0.43 & 1.00 & 1.00  \\ 
  \hline
\end{tabular}
\end{table}

   {\it On performance of BIDIFAC}: BIDIFAC was competitive against all compared models in separating signals into multiple shared structures. In Design 1 where $||\bG_{ij}||_F^2=||\bR_{ij}||_F^2=0$ 
    , the loss from the model misspecification was negligible when comparing the prediction errors for column-shared matrices ($\bC$) and the sum of global and column-shared matrices ($\bG+\bC$), as well as individual matrices ($\bI$) and the sum of row and individual matrices ($\bR+\bI$). The negligible effect of misspecification can also be seen when compared to UNIFAC, where the global or row-shared structures are ignored. In Design 2, where BIDIFAC is the only method estimating global ($\bG$) and row-shared ($\bR$) components, the prediction errors were comparable to the other components regardless of SNR. Even when error variances differ by each matrix, BIDIFAC was not affected severely.
    
    {\it On soft and hard thresholding}: When SNR is low, matrix completion via soft thresholding generally outperformed hard-thesholding approaches (JIVE, SLIDE and SVD(T)). It is easily seen by comparing SVD(soft) to SVD(T) or comparing JIVE(T)/JIVE(P)/SLIDE to UNIFAC. Especially, { when SNR$=0.5$ in} Design 1, JIVE(T) overfitted severely even when true ranks are given. JIVE(P) performed even worse, due to erroneous rank selection. { SLIDE suffered less from overfitting, but it was because it selected 0 rank in most simulated data.} As SNR increases, SLIDE, JIVE(P) and JIVE(T) outperformed UNIFAC and BIDIFAC in estimating $\bG+\bC$ and $\bR+\bI$ in both designs. This result is intuitive: soft-thresholding prevents over-fitting when the SNR is low, but over-penalizes the estimated signal when SNR is high. {We found that AJIVE did not perform well unless SNR is 2, where it provides provides similar results to JIVE with the true ranks, JIVE(T)}. 
    
     {\it On overall signal recovery}: In Design 1, JIVE(T) can be considered as the gold standard for overall signal recovery as it reflects the true joint and individual rank structures, which are unknown in practice. Except when SNR is 0.5 (which is explained by the difference between soft and hard thresholding), JIVE(T), JIVE(P) and SVD(soft) performed better than BIDIFAC in recovering the overall signals. The overall signal recovery of BIDIFAC was the same as UNIFAC and better than SVD(soft), revealing that BIDIFAC and UNIFAC obtained additional power from the column-shared matrices and the effect of model misspecification of BIDIFAC is negligible. In Design 2, BIDIFAC performed better than UNIFAC and SVD(soft) as it closely matched the data generating process.
    
    {\it On imputation performance}: JIVE(T) was the winner in both designs in imputing missing cells and columns except when SNR is 0.5.  However, recall that JIVE(T) uses the true ranks which are unknown in practice. BIDIFAC performed close to or even better than JIVE(P) in both designs, revealing that signal detection does not necessarily guarantee imputation performance as it is also affected by appropriate rank selection and detecting shared structures. In Design 1, it is not surprising that all models suffered when an entire row is missing, as there was no row-shared structure. In Design 2, BIDIFAC was the only method that successfully imputes missing rows.

To summarize, the performance of BIDIFAC was promising in simulation studies even when it was misspecified. It appropriately separated signals into linked structures and did not overfit for low SNR. Among the compared models, BIDIFAC is the only model that well accommodates the cases where a whole data matrix is missing or both a whole column and a whole row is missing. Acknowledging that it is not possible to obtain the true ranks of shared and individual structures, BIDIFAC performed the best across different scenarios.

\section{Data Analysis}
\label{sec:real}
\subsection{TCGA Breast Cancer Data}

In this section, we apply our method to breast cancer data from TCGA\citep{tcga12}. We integrate mRNA and miRNA profiles of the tumor samples and NATs, which, to our knowledge, has not been previously investigated in a fully unified framework. The data used here are freely available from TCGA. Specifically, we obtained the level III raw count of mRNA (RNASeq-V2) and miRNA (miRNA-Seq) data using the R package \texttt{TCGA2STAT} \citep{tcga2stat}. In our analysis we first removed the tumor data of those samples with matched NAT, so that tumor and NAT data correspond to two independent cohorts. We also filtered mRNAs and miRNAs with more than half zero counts for all individuals. 
Then we took $\log (1+\mbox{count})$ and centered each of mRNA and miRNA profiles to the mean of tumors and NATs, so that the mean of each row of each matrix is 0. {Lastly, we selected the 500 mRNAs and miRNAs with maximum variability.} The resulting data had $500$ mRNA and miRNA profiles for $660$ tumor samples and $86$ NATs.

Breast cancer tumor samples are classified into 5 intrinsic subtypes, based on expression levels of 50 pre-defined genes: Luminal A (LumA), Luminal B (LumB), HER2-enriched (HER2), Basal-enriched (Basal), and Normal-like tumors \citep{ciriello15}. In our TCGA data, 419 out of 660 tumor samples had the labeled subtypes. 

\subsection{Results}

We first applied BIDIFAC to the processed data until convergence. We summarized the proportion of explained variance, { as well as the estimated ranks of the components}, in Table \ref{tbl:3}. The difference between $R_{\bX_{ij}}^2(\bS_{ij})-R_{\bX_{ij}}^2(\bG_{ij}+\bR_{ij}+\bC_{ij})$ among tumors and NATs suggests that most of the variability of the miRNA and mRNA profiles  were attributed to shared structures. In particular, more than a half of the variability of both the mRNA and miRNA profiles from the NAT is attributed to the row-shared components (Global+Row); this makes sense, as tumor cells are derived from normal tissue and thus we expect much of the same patterns of variability that are present in normal tissue to also be present in tumors. Between {30-45\%} of the variability is explained by the shared structure between miRNA and mRNA across the four matrices. There is a large difference of sample sizes between tumor and normal tissues, which may have also affected the estimate of proportion of variance explained. {Even though the rank of the estimated signals in NAT was close to the rank of the data matrix (86) linear independence among the terms of the decomposition was preserved in our decomposition.}  

\begin{table}[!ht]
\centering
\caption{Proportion of variance explained (Equation \eqref{eq:0}) for each component, estimated by BIDIFAC using the breast cancer data. The parentheses denote the rank of the estimated components.}\label{tbl:3}
\begin{tabular}{rrrrrr}
  \hline
 & Global & Global+Row & Global+Col & Global+Row+Col & Signal \\
  \hline
Tumor mRNA & 0.14 (34) & 0.32 (68) & 0.45 (93) & 0.58 (127) & 0.67 (173)\\ 
  NAT mRNA & 0.23 (34) & 0.50 (68) & 0.44 (41) & 0.66 (75) & 0.78 (83) \\ 
  Tumor miRNA & 0.09 (34) & 0.46 (67) & 0.30 (93) & 0.63 (126) & 0.76 (175) \\ 
  NAT miRNA & 0.13 (34) & 0.66 (67) & 0.24 (41) & 0.75 (74) & 0.76 (79)\\ 
   \hline
\end{tabular}
\end{table}

{We compared BIDIFAC to existing methods on one-dimensionally linked matrices, including UNIFAC, JIVE(P) and SLIDE. We used two criteria to compare models: subtype classification and survival analysis. Specifically, we hypothesize that accounting for multiple –omics profiles and removing the shared variations between tumors and normal tissues would increase the biological interpretation of cancer subtypes and patient's survival. Thus, we focused on column-shared and individual components estimated by BIDIFAC and compared it to other methods using tumor mRNA and miRNA only.}

\begin{figure}[!ht]
    \includegraphics[scale=0.8]{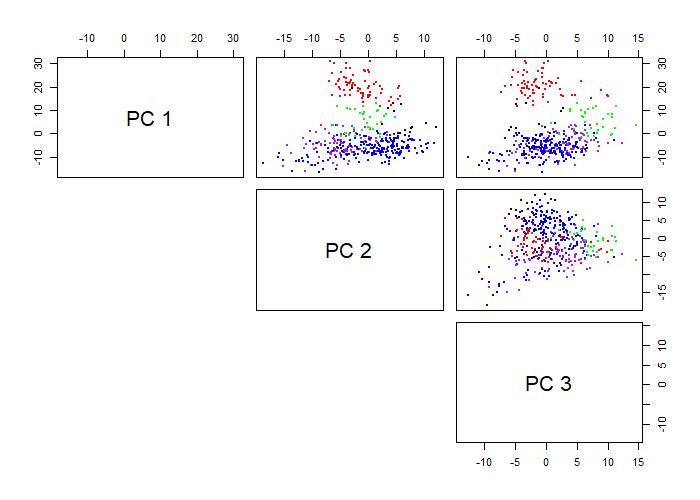}
    \caption{Visualizations of 5 breast cancer subtypes using the top three principal components of the column-shared component from BIDIFAC.} 
    \label{fig:3}
\end{figure}

We first compared how the estimated components are well distinguished by breast cancer subtypes. To summarize the subtype distinctions, we used the SWISS (Standardized WithIn Class Sum of Squares) score \citep{cabanski10}. Interpreted similar to ANOVA, a lower SWISS score implies more clear subtype distinction. We restricted the attention to the tumor samples with labeled subtypes. {As summarized in Table \ref{tbl:4}, the SWISS score of $\bC+\bI$ of UNIFAC was superior to JIVE and SLIDE, suggesting that soft-thresholding better uncovered subtype heterogeneity when integrating mRNA and miRNA. Compared to UNIFAC, BIDIFAC slightly obtained better SWISS score with lower rank of the $\bC+\bI$. The scatterplot of the principal components from the column-shared structures of BIDFAC, shown in Figure \ref{fig:3}, reveals that the subtypes are well distinguished. The column-shared structure from JIVE(P) had the lowest SWISS score overall, which may be because of its extremely low rank.} 

\begin{table}{!ht}
\caption{Summary of the SWISS scores and $p$-values of the score test of the estimated components from the Cox proportional hazards model, including tumors only. `Rank' refers to the rank of the estimated components.}\label{tbl:4}
\centering
\begin{tabular}{rlrrr}
  \hline
Model &  Components & Rank & SWISS & $p$-value  \\
  \hline
BIDIFAC & Signal & 173 & 0.54& 0.002   \\
        & Global & 34 & 0.69& 0.046 \\
        & Row & 34 & 0.75 & 0.085 \\
         & Col+Indiv & 105 & 0.52& 0.003 \\
        & Col & 59 & 0.48&  0.029 \\ 
        &  Indiv & 46 & 0.79& 0.003\\ \hline
UNIFAC  & Col+Indiv & 137& 0.54 & 0.001\\
        &  Col & 75& 0.49 & 0.012\\
        &  Indiv & 62 & 0.73& 0.001 \\ \hline
JIVE(P) & Col+Indiv & 35 & 0.64  & 0.020 \\
        & Col & 3 &  0.31 &  0.007 \\
        &  Indiv & 32 & 0.90    & 0.114 \\ \hline
SLIDE   & Col+Indiv & 58 & 0.67  & 0.002 \\
        & Col & 16 & 0.52 &  0.007 \\
        &  Indiv & 42 & 0.96    & 0.061 \\ \hline
\end{tabular}
\end{table}

{Using overall survival data for patients with tumors, we applied the Cox proportional hazards model (PHM) using scores from the estimated components as predictors and used the score test to assess its significance. The results are also shown in Table \ref{tbl:4}. At $\alpha=0.05$, all models suggested that scores from $\bC+\bI$ were associated. Narrowing down the scope to each structural component, we also found that the individual components of UNIFAC and BIDIFAC, even though not indicative of subtype distinction in SWISS scores, could provide additional information on patients' survival. BIDIFAC, accounting for normal tissues, provided reduced rank for individual components compared to UNIFAC, which would contribute to the increased power of the test.}

\section{Conclusion}
\label{sec:conc}
In this paper we propose BIDIFAC, the first unified framework to handle multiple matrices that are shared both vertically and horizontally. In contrast to existing methods on joint matrix factorization, we provide the estimator based on soft-thresholding and nuclear norm penalization, where the complicated problem of selecting tuning parameters is alleviated by using the well-known result from random matrix theory. We conducted extensive simulation studies to show the efficiency and flexibility of BIDIFAC compared to existing methods.  We applied our method to TCGA breast cancer data, where mRNA and miRNA profiles are obtained separately from both tumor tissues and normal tissues adjacent to tumors (NATs). 
{From this application we conclude that (i) patterns of variability in normal tissue are largely also present in tumor tissue for both mRNA and miRNA, (ii) patterns that are associated with survival and clinical subtypes across both mRNA and miRNA are largely *not* present in normal tissue, and (iii) patterns that distinguish the clinical subtypes are shared by mRNA and miRNA. Existing vertical integration methods establish (iii), but not (i) or (ii).}     
In addition to the integration of tumor and NAT data, this methodology may be applied to a growing number of applications with bidimensionally linked matrices.  A particularly intriguing potential application is the integration of ``pan-omics pan-cancer" data; that is, the integration of multi-omic data for samples from multiple types of cancers (as defined by their tissue-of-origin).

We describe several limitations of our method. These limitation primarily relate to use of the Frobenious norm in the objective and applying random matrix theory to select tuning parameters. Importantly, our model assumes normality of each matrix to select appropriate tuning parameters, which may be violated in practice.  For example, application to SNPs data would be limited as each cell of a matrix takes discrete values (e.g., $0, 1, 2$). Similarly, we do not consider the case where one or more matrices are binary, which is common for biomedical data. Also, our approach is sensitive to outliers, due to the use of the Frobenious norm in the objective function. Even when the Gaussian assumption holds, our model may be extended by adopting variable sparsity or overcoming deficiencies of soft-thresholding, as neither soft-thresholding nor hard-thresholding provides optimal signal recovery \citep{shabalin13}. { Although our convex objective guarantees convergence to a { stationary point} that is empirically consistent for different starting values, the theoretical identifiability properties of the resulting decomposition deserves further study.}  Lastly, {our approach does not inherently model uncertainty in the underlying structural decomposition and missing value imputations; the result gives the mode of a Bayesian posterior, and extensions to fully Bayesian approaches are worth considering.}      

Here we have focused on capturing four types of low-rank signals in bidimensional data: global, column-shared, row-shared, and individual.   Other joint signals are possible.  For example, when $p=3,q=1$, it is possible that $\bX_{11}$ and $\bX_{21}$ share signal that is not present in $\bX_{31}$. Similarly, our method would suffer when $\bX_{11},\bX_{12},\bX_{21}$ share signal that is not present in $\bX_{22}$, though this is perhaps less likely for most data applications.   It is straightforward to extend our framework to accommodate these or other structures, with appropriate additions to the objective \eqref{eq:als}.  

An interesting extension of our work is the factorization of higher-order arrays (i.e., tensors). For example, integrating multiple {\it omics} profiles for multiple cohorts that are measured at multiple time points would give additional insights on the linked structure in a longitudinal manner that cross-sectional designs cannot provide.

\section{Availability}
\label{sec:availability}
We provide R functions for fitting BIDIFAC, imputing missing values, and generating simulation data as in Section \ref{sec:sims}, which are available at (\href{https://github.com/lockEF/bidifac}{\texttt{https://github.com/lockEF/bidifac}}).

\section*{Acknowledgement}
{ We would like to thank co-editor, associate editor, and two anonymous reviewers for their constructive comments. }
This research was supported by the National Institute of Health (NIH) under the grant R21CA231214 and by the Minnesota Supercomputing Institute (MSI).

\appendix 

\section{Proposed Algorithm (ISSVT)}\label{algorithm}
The ISSVT algorithm, proceeds as follows proceeds as follows:
\begin{enumerate}
    \item Initialize $ \widehat{\bThet}=\lbrace \widehat{\bG}_{00}, \widehat{\bR}_{i0}, \widehat{\bC}_{0j},\widehat{\bI}_{ij}~|~i=1,\dots, p, j=1,\dots, q\rbrace$.
    \item For $i=1,\dots,p$ and $j=1,\dots,q$, apply Proposition 1 to obtain a closed form solution of the following:
    \begin{enumerate}
        \item $\widehat{\bG}_{00}^{(new)}=\underset{\bG_{00}:r(\bG_{00})\leq r_{00}}{\arg\min} \dfrac{1}{2}||\bX_{00}-\widehat{\bR}_{00}-\widehat{\bC}_{00}-\widehat{\bI}_{00}-\bG_{00}||_F^2+\lambda_{00}||\bG_{00}||_*$
        
        \item $\widehat{\bR}_{i0}^{(new)}=\underset{\bR_{i0}:r(\bR_{i0})\leq r_{i0}}{\arg\min}\dfrac{1}{2}||\bX_{i0}-\widehat{\bG}_{i0}^{(new)}-\widehat{\bC}_{i0}-\widehat{\bI}_{i0}-\bR_{i0}||_F^2+\lambda_{i0} ||\bR_{i0}||_*$
        
        \item $\widehat{\bC}_{0j}^{(new)}=\underset{\bC_{0j}:r(\bC_{0j})\leq r_{0j}}{\arg\min}\dfrac{1}{2}||\bX_{0j}-\widehat{\bG}_{0j}^{(new)}-\widehat{\bR}_{0j}^{(new)}-\widehat{\bI}_{0j}-\bC_{0j}||_F^2+\lambda_{0j} ||\bC_{0j}||_*$
        \item $\widehat{\bI}_{ij}^{(new)}=\underset{\bI_{ij}:r(\bI_{ij})\leq r_{ij}}{\arg\min}\dfrac{1}{2}||\bX_{ij}-\widehat{\bG}_{ij}^{(new)}-\widehat{\bR}_{ij}^{(new)}-\widehat{\bC}_{ij}^{(new)}-\bI_{ij}||_F^2+\lambda_{ij} ||\bI_{ij}||_*$
    \end{enumerate}
    \item The algorithm converges if $f_2(\widehat{\bThet})- f_2(\widehat{\bThet}^{(new)})< \epsilon$.
    If it does not converge, replace $\widehat{\bThet}^{(new)}$ with $\widehat{\bThet}$ and repeat Step 2.
    
\end{enumerate}
The algorithm iteratively minimizes the objective $f_2$ (6) over blocks $\bG_{00}$, $\{\bR_{i0} \mid i=1,\hdots,p\}$, $\{\bC_{0j}  \mid j=1,\hdots,q\}$, and $ \{\bI_{ij} \mid i=1,\dots, p, j=1,\dots, q\}$.   By Proposition 1, this is equivalent to a blockwise coordinate descent algorithm for $f_1$ (4), with corresponding update blocks $\{\bU_{i0}^{(G)}, \bV_{0j}^{(G)} \mid i=1,\dots, p, j=1,\dots, q\}$, $\{\bU_{ij}^{(C)}, \bV_{0j}^{(C)} \mid i=1,\dots, p, j=1,\dots, q\}$, $\{\bU_{i0}^{(R)}, \bV_{ij}^{(R)} \mid i=1,\dots, p, j=1,\dots, q\}$, $\{\bU_{ij}^{(R)}, \bV_{ij}^{(I)} \mid i=1,\dots, p, j=1,\dots, q\}$.

\section{Proofs}\label{proofs}

 \begingroup
\def\thetheorem{\ref{thm:1}}
\begin{theorem}
Let \[\widehat{\bThet}_1=\lbrace\widehat{\bU}_{i0}^{(G)}\widehat{\bV}_{0j}^{(G)T}, \widehat{\bU}_{i0}^{(R)}\widehat{\bV}_{ij}^{(R)T}, \widehat{\bU}_{ij}^{(C)}\widehat{\bV}_{0j}^{(\bC)T}, \widehat{\bU}_{ij}^{(I)}\widehat{\bV}_{ij}^{(I)T} \mid i=1\ldots p, j=1,\ldots q\rbrace\] minimize (4).  Then, \[\widehat{\bThet}_2 = \lbrace \widehat{\bG}_{ij}, \widehat{\bR}_{ij}, \widehat{\bC}_{ij}, \widehat{\bI}_{ij} \mid i=1\ldots p, j=1,\ldots q \rbrace\] minimizes (6), where  $\widehat{\bG}_{ij}=\widehat{\bU}_{i0}^{(G)}\widehat{\bV}_{0j}^{(G)T}, \widehat{\bR}_{ij}=\widehat{\bU}_{i0}^{(R)}\widehat{\bV}_{ij}^{(R)T}, \widehat{\bC}_{ij}=\widehat{\bU}_{ij}^{(C)}\widehat{\bV}_{0j}^{(C)T},$ and  $\widehat{\bI}_{ij}=\widehat{\bU}_{ij}^{(I)}\widehat{\bV}_{ij}^{(I)T}$.  
\end{theorem}
\addtocounter{theorem}{-1}
\endgroup

\begin{proof}
Because $\widehat{\bThet}_1$ minimizes $f_1$,  \[||\widehat{\bU}_{ij}^{(I)}||_F^2 + ||\widehat{\bV}_{ij}^{(I)}||_F^2 = \underset{\substack{\bU\bV^T=\widehat{\bI}_{ij}\\ \bU: m_1\times \min(m_1,n_1)\\ \bV:n_1\times \min(m_1,n_1)}}{\min}(||\bU||_F^2+||\bV||_F^2).\]
It follows from Lemma 1 that 
\[||\widehat{\bU}_{ij}^{(I)}||_F^2 + ||\widehat{\bV}_{ij}^{(I)}||_F^2 = 2 ||\widehat{\bI}_{ij}||_*  \;  \text{ for all } i>0, j>0.\]
Analogous arguments show that 
\begin{align*}
||\widehat{\bU}_{i0}^{(R)}||_F^2+||\widehat{\bV}_{i0}^{(R)}||_F^2 &= 2 ||\widehat{\bR}_{i0}||_* \; \text{ for all } i>0 \\  
||\widehat{\bU}_{0j}^{(C)}||_F^2+||\widehat{\bV}_{0j}^{(C)}||_F^2 &= 2 ||\widehat{\bC}_{0j}||_* \; \text{ for all } j>0, \; \text{and} \\
||\widehat{\bU}_{00}^{(G)}||_F^2+||\widehat{\bV}_{00}^{(G)}||_F^2 &= 2 ||\widehat{\bG}_{00}||_*.
\end{align*}
Thus, $f_1(\widehat{\bThet}_1)=2 f_2(\widehat{\bThet}_2)$.

Consider an alternative estimate $\widetilde{\bThet}_2 = \lbrace \widetilde{\bG}_{ij}, \widetilde{\bR}_{ij}, \widetilde{\bC}_{ij}, \widetilde{\bI}_{ij} \mid i=1\ldots p, j=1,\ldots q \rbrace$.  By Lemma 1, $2 f_2(\widetilde{\bThet}_2)= f_1(\widetilde{\bThet}_1)$ for some \[\widetilde{\bThet}_1=\lbrace\widetilde{\bU}_{i0}^{(G)}\widetilde{\bV}_{0j}^{(G)T}, \widetilde{\bU}_{i0}^{(R)}\widetilde{\bV}_{ij}^{(R)T}, \widetilde{\bU}_{ij}^{(C)}\widetilde{\bV}_{0j}^{(C)T}, \widetilde{\bU}_{ij}^{(I)}\widetilde{\bV}_{ij}^{(I)T} \mid i=1\ldots p, j=1,\ldots q\rbrace\] where $\widetilde{\bG}_{ij}=\widetilde{\bU}_{i0}^{(G)}\widetilde{\bV}_{0j}^{(G)T}, \widetilde{\bR}_{ij}=\widetilde{\bU}_{i0}^{(R)}\widetilde{\bV}_{ij}^{(R)T}, \widetilde{\bC}_{ij}=\widetilde{\bU}_{ij}^{(C)}\widetilde{\bV}_{0j}^{(C)T},$ and  $\widetilde{\bI}_{ij}=\widetilde{\bU}_{ij}^{(I)}\widetilde{\bV}_{ij}^{(I)T}$. Thus, because $\widehat{\bThet}_1$ minimizes $f_1$,  
\[2 f_2(\widetilde{\bThet}_2)= f_1(\widetilde{\bThet}_1) \geq f_1(\widehat{\bThet}_1) = 2 f_2(\widehat{\bThet}_2),\]
and we conclude that $\widehat{\bThet}_2$ minimizes $f_2$.  

\end{proof}

 \begingroup
\def\thetheorem{\ref{thm:2}}
\begin{theorem}
The objective $f_2(\cdot)$ in (6) is convex over its domain.  
\end{theorem}
\addtocounter{theorem}{-1}
\endgroup

\begin{proof}
Consider $\widetilde{\bThet}^{(m)} = \lbrace \bG_{ij}^{(m)}, \bR_{ij}^{(m)}, \bC_{ij}^{(m)}, \bI_{ij}^{(m)} \mid  i=1,\ldots, p, j=1,\ldots q \rbrace$ for $m=1,2$ and $\alpha \in [0,1]$.  It suffices to show \begin{align}f_2 \left(\alpha \widetilde{\bThet}^{(1)} + (1-\alpha) \widetilde{\bThet}^{(2)} \right) \leq \alpha f_2(\widetilde{\bThet}^{(1)}) + (1-\alpha) f_2(\widetilde{\bThet}^{(2)}), \label{result} \end{align}
where $\alpha \widetilde{\bThet}^{(1)} + (1-\alpha) \widetilde{\bThet}^{(2)}$ is given by  
\begin{align*}
    \lbrace \alpha \bG_{ij}^{(1)}+(1-\alpha)\bG_{ij}^{(2)}, \alpha \bR_{ij}^{(1)}+(1-\alpha)\bR_{ij}^{(2)}, \alpha \bC_{ij}^{(1)}+(1-\alpha)&\bC_{ij}^{(2)}, \alpha \bI_{ij}^{(1)}+(1-\alpha)\bI_{ij}^{(2)} \\ & \mid i=1,\ldots, p, j=1,\ldots q \rbrace.
\end{align*}
Decompose $f_2(\bThet) = f_2^{\text{LS}}(\bThet) + f_2^{\text{PEN}}(\bThet)$, where
\begin{align*}
f_2^{\text{LS}}(\bThet) &= \frac{1}{2}\sum_{i=1}^p\sum_{j=1}^q ||\bX_{ij}-\bG_{ij}-\bR_{ij}-\bC_{ij}-\bI_{ij}||_F^2, \; \; \text{and} \\
f_2^{\text{PEN}}(\bThet) &= \lambda_{00}||\bG_{00}||_*+\sum_{i=1}^p \lambda_{i0}||\bR_{i0}||_*+\sum_{j=1}^q \lambda_{0j}||\bC_{0j}||_*+\sum_{i=1}^p\sum_{j=1}^q \lambda_{ij}||\bI_{ij}||_*.
\end{align*}
By convexity of the least squares objective 
\begin{align}||\bX_{ij}- (\alpha \bA^{(1)}_{ij} +(1-\alpha) \bA^{(2)}_{ij})||_F^2 \leq \alpha ||\bX_{ij}-  \bA^{(1)}_{ij}||_F^2 + (1-\alpha) ||\bX_{ij}-  \bA^{(2)}_{ij}||_F^2 \label{ls_conv}  \end{align}
for any $\bA_{ij}^{(1)}$ and $\bA_{ij}^{(2)}$.  Applying~\eqref{ls_conv} for each $(i,j)$, where  $\bA^{(m)}_{ij} = \bG^{(m)}_{ij}+\bR^{(m)}_{ij}+\bC^{(m)}_{ij}+\bI^{(m)}_{ij}$ for $m=1,2$, gives 
\begin{align}
f_2^{\text{LS}}\left(\alpha \widetilde{\bThet}^{(1)} + (1-\alpha) \widetilde{\bThet}^{(2)} \right) \leq \alpha f^{\text{LS}}_2(\widetilde{\bThet}^{(1)}) + (1-\alpha) f^{\text{LS}}_2(\widetilde{\bThet}^{(2)}).
\label{ls_conv2}    
\end{align}
By convexity of the nuclear norm operator,
\begin{align}||\alpha \bA^{(1)} + (1-\alpha) \bA^{(2)}||_* \leq \alpha ||\bA^{(1)}||_* + (1-\alpha)||\bA^{(2)}||_* \label{pen_conv} \end{align}
for any $\bA^{(1)}$ and $\bA^{(2)}$. Applying~\eqref{pen_conv} to each additive term in $f_2^{\text{PEN}}$ gives 
\begin{align}
 f_2^{\text{PEN}}\left(\alpha \widetilde{\bThet}^{(1)} + (1-\alpha) \widetilde{\bThet}^{(2)} \right) \leq \alpha f^{\text{PEN}}_2(\widetilde{\bThet}^{(1)}) + (1-\alpha) f^{\text{PEN}}_2(\widetilde{\bThet}^{(2)}).
 \label{pen_conv2}
 \end{align}
 Thus,~\eqref{ls_conv2} and~\eqref{pen_conv2} imply~\eqref{result}.
\end{proof}

 \begingroup
\def\theprop{\ref{prop3}}
\begin{prop}
The following conditions are necessary to allow for non-zero $\widehat{\bG}_{00}$, $\widehat{\bR}_{i0}$, $\widehat{\bC}_{0j}$, and $\widehat{\bI}_{ij}$: 
   \begin{enumerate}
   \item $ \mbox{max}_j \lambda_{ij} < \lambda_{i0} < \sum_j  \lambda_{ij}$ for $i=1,\hdots,p$
   and $\mbox{max}_i \lambda_{ij} < \lambda_{0j} < \sum_i  \lambda_{ij} $ for $j=1,\hdots,q$
   \item $\mbox{max}_i \lambda_{i0} < \lambda_{00} < \sum_i  \lambda_{i0} $
   and $\mbox{max}_j \lambda_{0j} < \lambda_{00} < \sum_j  \lambda_{0j} $.
   \end{enumerate}
\end{prop}
\addtocounter{prop}{-1}
\endgroup

\begin{proof}
Consider a violation of the left-hand inequality in condition 1.: $\lambda_{ij}\geq \lambda_{i0}$.  Define $\widehat{\bThet}$ to be identical to $\widetilde{\bThet}$ but with $\widehat{\bI}_{ij}=\mathbf{0}$ and $\widehat{\bR}_{ij}=\widetilde{\bR}_{ij}+\widetilde{\bI}_{ij}$.  By convexity of the nuclear norm, $||\widehat{\bR}_{i0}||_* \leq  ||\widetilde{\bR}_{i0}||_*+||\widetilde{\bI}_{ij}||_*$, and it follows that 
\begin{align*}
f_2(\widetilde{\bThet})-f_2(\widehat{\bThet}) &= \lambda_{i0} ||\widetilde{\bR}_{i0}||_*+\lambda_{ij}||\widetilde{\bI}_{ij}||_*-\lambda_{i0}||\widehat{\bR}_{i0}||_* \\ 
&\geq \lambda_{i0} ||\widetilde{\bR}_{i0}||_*+\lambda_{ij}||\widetilde{\bI}_{ij}||_*-\lambda_{i0} (||\widetilde{\bR}_{i0}||_*+||\widetilde{\bI}_{ij}||_*)   \\
& \geq \lambda_{i0} ||\widetilde{\bR}_{i0}||_*+\lambda_{ij}||\widetilde{\bI}_{ij}||_*-\lambda_{i0} ||\widetilde{\bR}_{i0}||_*-\lambda_{ij}||\widetilde{\bI}_{ij}||_* \\
&=0. 
\end{align*}
Thus, regardless of the data $\bX_{00}$, the objective $f_2(\bThet)$ is minimized with $\widehat{\bI}_{ij}=\bzero_{m_i\times n_j}$.  An analogous argument show that a violation of $\lambda_{ij}<\lambda_{0j}$ implies $\widehat{\bI}_{ij}=\bzero_{m_i\times n_j}$ for some $i,j$. Moreover, analogous arguments show that a violation of $\max_i\lambda_{i0}<\lambda_{00}$ implies $\widehat{\bR}_{i0}=\bzero_{m_i\times n_0}$ for some $i$, and that a violation of $\max_j \lambda_{0j}<\lambda_{00}$ implies $\widehat{\bC}_{0j}=\bzero_{m_0\times n_j}$ for some $j$. 

Now, consider a violation of the right-hand inequality of condition 1.: $\lambda_{i0}\geq \sum_j \lambda_{ij}$.  Define $\widehat{\bThet}$ to be identical to $\widetilde{\bThet}$ but with $\widehat{\bR}_{i0}=\bzero_{m_i\times n_0}$ and $\widehat{\bI}_{ij}=\widetilde{\bR}_{ij}+\widetilde{\bI}_{ij}$ for $j=1,\hdots,q$.  Then,  
\begin{align*}
f_2(\widetilde{\bThet})-f_2(\widehat{\bThet}) &= \lambda_{i0} ||\widetilde{\bR}_{i0}||_*+\sum_j \lambda_{ij}||\widetilde{\bI}_{ij}||_*-\sum_j \lambda_{ij}||\widehat{\bI}_{ij}||_* \\ 
&\geq \lambda_{i0} ||\widetilde{\bR}_{i0}||_*+\sum_j \lambda_{ij}||\widetilde{\bI}_{ij}||_*-\sum_j \lambda_{ij} (||\widetilde{\bI}_{ij}||_*+||\widetilde{\bR}_{ij}||_*)    \\
& = \lambda_{i0} ||\widetilde{\bR}_{i0}||_* - \sum_j \lambda_{ij} ||\widetilde{\bR}_{ij}||_* \\
&\geq \lambda_{i0} ||\widetilde{\bR}_{i0}||_* - \sum_j \lambda_{ij} ||\widetilde{\bR}_{i0}||_* \\
&\geq 0. 
\end{align*} 
Thus, regardless of the data $\bX_{00}$, the objective $f_2(\bThet)$ is minimized with $\widehat{\bR}_{i0}=\bzero_{m_i\times n_0}$. An analogous argument show that a violation of $\sum_{0j}<\sum_i\lambda_{ij}$ implies $\widehat{\bC}_{0j}=\bzero_{m_0\times n_j}$ for some $j$.  Moreover, analogous arguments show that a violation of $\lambda_{00}<\sum_i \lambda_{i0}$ or $\lambda_{00}<\sum_j\lambda_{0j}$  imply $\widehat{\bG}_{00}=\bzero_{m_0\times n_0}$. \hfill $\square$
\end{proof}

\section{Noiseless simulation}\label{NoNoiseSim}

Here we generate joint and individual signals under the UNIFAC (vertically linked) model, and assess their recovery without residual error.  We contrast the decomposition provided by the penalized objective  $f_2(\cdot)$, with that obtained by alternative methods that enforce orthogonality of the estimated components. 

We generate $\mathbf{X}_1: d_1 \times n$ and $\mathbf{X}_2: d_2 \times n$ as a sum of low-rank column-shared and individual structures that are \emph{independent} but not necessarily \emph{orthogonal}.  That is,  
\begin{align*}
\mathbf{X}_{1} &= \mathbf{C}_1+\mathbf{I}_1\\
\mathbf{X}_{2} &= \mathbf{C}_2+\mathbf{I}_2
\end{align*}
where 
\[\bC_i = \mathbf{U}_i^{(C)} \mathbf{V}^T \;  \text{ and } \; \bI_i = \mathbf{U}_i^{(I)} \mathbf{V}_i^T  \;  \text{ for } \; i=1,2,\]
and the entries of $\mathbf{U}_1^{(C)}: d_1 \times r,\mathbf{U}_2^{(C)}: d_1 \times r, \mathbf{V}: n \times r, \mathbf{U}_1^{(I)}: d_1 \times r_1,\mathbf{V}_1: n \times r_2, \mathbf{U}_2^{(I)}: d_2 \times r_2, \mathbf{V}_2: n \times r_2$ are generated independently from a $\mathcal{N}(0,1)$ distribution.  

We fix $r=r_1=r_2=10$ and generate $10$ datasets under each of four scenarios with different row and column dimensions: (1) $d_1=d_2=n=100$, (2) $d_1=d_2=100, n=500$, (3) $d_1=d_2=500, n=100$, (4) $d_1=d_2=n=500$.  For each generated dataset we apply (i) UNIFAC, (ii) JIVE, (iii) AJIVE, and (iv) SLIDE, where the correctly specified ranks are used for methods (ii--iv).  For UNIFAC, the noise variance is set to a small value ($\sigma=0.0001$) and the default penalties are used.  In each case we compute the mean relative error in recovering underlying joint and individual structures:
\begin{align}\mbox{PredErr}(\widehat{\mathbf{C}}, \widehat{\mathbf{I}}) = \frac{1}{4} \left(  \frac{||\mathbf{C}_1-\widehat{\mathbf{C}}_1||_F^2}{||\mathbf{C}_1||_F^2}+\frac{||\mathbf{I}_1-\widehat{\mathbf{I}}_1||_F^2}{||\mathbf{I}_1||_F^2} + \frac{||\mathbf{C}_2-\widehat{\mathbf{C}}_2||_F^2}{||\mathbf{C}_2||_F^2}+\frac{||\mathbf{I}_2-\widehat{\mathbf{I}}_2||_F^2}{||\mathbf{I}_2||_F^2}\right). \label{meanrelerror} \end{align}
The results are summarized in Table~\ref{tab:nonoise}. All methods decompose the underlying joint and individual signals with negligible error as the dimension of the sample size ($n$) and dimensions ($d=d_2=d_2$) increase.  UNIFAC recovers the joint and individual signals with comparable or substantially improved accuracy across the four scenarios, despite the use of the correct ranks for the other three methods.  The error in recovery for methods (ii-iv) is due to the orthogonality constraints, which are necessary for identifiability of the decomposition without additional penalization. The independent joint and individual structures are not exactly orthogonal, but will approach orthogonality as $n\rightarrow \infty$, and thus performance improves for $n=500$ vs $n=100$.    JIVE and AJIVE both assume orthogonality of the rows of the joint and individual structure, $\bC$ and $\bI$; thus, their performance is comparable across scenarios and differences are solely due to imprecision of the computational algorithm for JIVE.  SLIDE additionally assumes orthogonality of the individual structures $\bI_1$ and $\bI_2$, which results in slightly less accurate recovery for lower $n$.     Even under independence the underlying joint and individual signals are not precisely orthogonal.  The performance of UNIFAC for higher dimension $d$ demonstrates the potential to recover the true joint and individual signals more accurately by relaxing orthogonality constraints.       

\begin{table}[!ht]
\centering
\caption{Mean relative recovery error for joint and individual signals, with standard error in parentheses, across different scenarios. }\label{tab:nonoise}
\begin{tabular}{l|c c c c}
  \hline
  & $d=100, n=100$ & $d=500, n=100$ & $d=100, n=500$ & $d=500, n=500$ \\ 
 \hline 
 UNIFAC & \textbf{0.058} (0.002) & \textbf{0.033} (0.001) & \textbf{0.027} (0.001)  & \textbf{0.010} (0.001) \\
 JIVE & \textbf{0.106} (0.003) & \textbf{0.110} (0.003) & \textbf{0.023} (0.001)  & \textbf{0.022} (0.001)\\
 AJIVE & \textbf{0.096} (0.003) & \textbf{0.100} (0.003) & \textbf{0.021} (0.001) & \textbf{0.021} (0.001)\\
 SLIDE & \textbf{0.132} (0.005) & \textbf{0.130} (0.004) & \textbf{0.026} (0.001) & \textbf{0.024} (0.001)\\ \hline
\end{tabular}
\end{table}

\begin{figure}[!ht]
    \centering
    \includegraphics[scale=1]{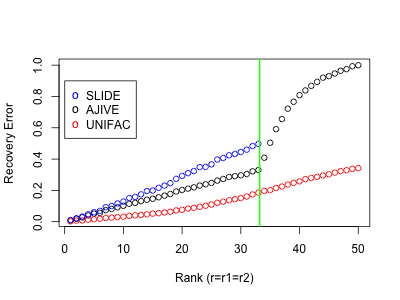}
    \caption{Mean relative recovery with data generated and estimated under the given ranks}
    \label{fig:nonoiserank}
 \end{figure}

We also compare the recovery of joint and individual signals as their ranks increase relative to the data dimensions.  We consider the scenario with  $d_1=d_2=500, n=100$, and generate $10$ datasets for each of $r_1=r_2=r=\{1,2,\hdots,50\}$.  For each dataset we estimate the decomposition via UNIFAC, AJIVE, or SLIDE; AJIVE is used instead of JIVE because in the noiseless scenario with given ranks they give the same underlying decomposition and AJIVE is more computationally efficient.  The resulting mean relative errors \eqref{meanrelerror} are shown in Figure~\ref{fig:nonoiserank}.  For ranks greater than $r=r_1=r_2=33$, the sum of the ranks of column-shared and individual structures ($r+r_1+r_2$) is greater than the rank of the observed signal: rank$([\mathbf{X}_{1}^T,\mathbf{X}_{2}^T ]) \leq 100$.  Thus, for these cases a SLIDE decomposition with the given ranks does not exist because the condition of orthogonality between $\hat{J}$, $\hat{I}_1$ and $\hat{I}_2$ cannot be satisfied.   The recovery errors for the AJIVE decomposition also increase sharply at this point, while the trend remains stable for UNIFAC.  In general, UNIFAC provides better recovery of the generated joint and individual signal as the ranks increase.  However, the recovery error does increase steadily as the ranks get larger.  Moreover, if the sum of the ranks is greater than the rank of the overall signal, this implies linear dependence among the underlying components (here, linear dependence among $\mathbf{V}$, $\mathbf{V}_1$, and $\mathbf{V}_2$), which complicates their interpretation.

 \section{Data Analysis: Residual Diagnostics}

\begin{figure}[!ht]
    \centering
    \includegraphics[scale=0.7]{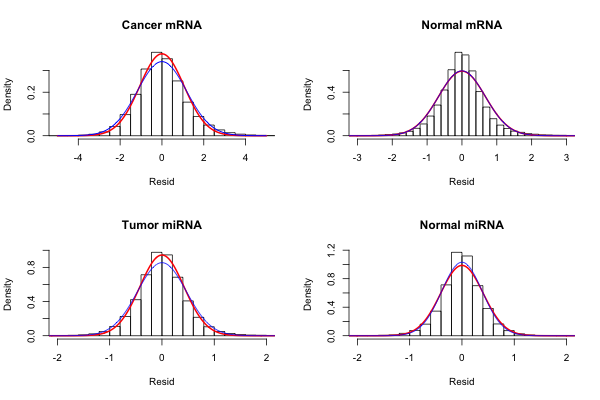}
    \caption{Distribution of residuals for the TCGA application.  The Gaussian density given by the noise variance estimate $\hat{\sigma}_{ij}^{MAD}$ is shown in \textcolor{red}{red}, the density given by the emperical mean and standard deviation is shown in \textcolor{blue}{blue}.}
    \label{fig:residplots}
 \end{figure}

Here we consider the distribution of residuals after the application of BIDIFAC for the TCGA application described in Section 4 of the main manuscript.  That is, we consider the residual matrices
\[\widehat{\bE}_{ij} = \widehat{\bX}_{ij}-\widehat{\bG}_{ij}-\widehat{\bC}_{ij}-\widehat{\bR}_{ij}-\widehat{\bI}_{ij}.\]
BIDIFAC is best motivated when the error terms $\bE_{ij}$ are approximately Gaussian, as the objective identifies the mode of a Bayesian model with a Gaussian likelihood (Section 2.7) and the selection of the tuning parameters is based on the assumption of Gaussian error (Section 2.6).  Figure~\ref{fig:residplots} shows the distribution of residuals for each of the four datasets considered (tumor mRNA, tumor miRNA, normal mRNA, normal miRNA) overlayed with two Gaussian densities: one giving the theoretical distribution of residuals implied by the pre-hoc estimate of the noise variance for each dataset ($\hat{\sigma}_{ij}^{MAD}$), and another giving the empirical Gaussian fit resulting from the sample mean and standard deviation of the observed residuals.  None of the residual histograms show strong departures from Gaussianity, and the variance estimate used to tune the model ($\hat{\sigma}_{ij}^{MAD}$) fits the observed residuals reasonably well in each case.

\end{document}